\pdfoutput=1

\documentclass[11pt]{article}

\usepackage[]{acl}

\usepackage{times}
\usepackage{latexsym}
\usepackage{amsmath}
\usepackage{graphicx}
\usepackage{amsfonts}
\usepackage{booktabs}
\usepackage{multirow}
\usepackage{xurl}

\usepackage[T1]{fontenc}

\usepackage[utf8]{inputenc}
\usepackage{scalerel,xparse}
\NewDocumentCommand\emojijelly{}{\includegraphics[scale=0.06]{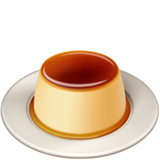} }

\NewDocumentCommand\emojijellybig{}{\includegraphics[scale=0.08]{EmojiFolder/custard.png}
}

\usepackage{microtype}

\title{Improving Dialogue Agents by Decomposing One Global Explicit Annotation with Local Implicit Multimodal Feedback}

\author{%
    Dong Won Lee$^1$ \quad
    Hae Won Park$^1$ \quad Yoon Kim$^1$ \quad Cynthia Breazeal$^1$ \quad Louis-Philippe Morency$^2$   \\
    Massachusetts Institute of Technology$^1$, Carnegie Mellon University$^2$\\
    \texttt{dongwonl@mit.edu}
}

\begin{document}
\maketitle
\begin{abstract}
 We describe an approach for  aligning an LLM-based dialogue agent based on global (i.e., dialogue-level) rewards, while also taking into account naturally-occurring multimodal signals. At a high level, our approach (dubbed  \textbf{GELI}) learns a local, turn-level reward model by decomposing the human-provided \textbf{G}lobal  \textbf{E}xplicit \textbf{(GE)} session-level reward, using \textbf{L}ocal \textbf{I}mplicit \textbf{(LI)} multimodal reward signals to crossmodally shape the reward decomposition step. This decomposed reward model is then used as part of the standard RLHF pipeline improve an LLM-based dialog agent. We run quantitative and qualitative human studies to evaluate the performance of our GELI approach, and find that it shows consistent improvements across various conversational metrics compared to baseline methods.

\end{abstract}

\section{Introduction}
Developing social dialogue agents that can interact and collaborate with humans over a long horizon remains a longstanding  goal of artificial intelligence. Large language models (LLM) pretrained at scale on the next-word prediction objective and then subsequently aligned to human preference via RLHF (Reinforcement with Human Feedback) represent a significant step in this direction \cite{ouyang2022training}, even leading to successful commercial applications.

However, existing methods for alignment usually assume that preference labels are annotated at the \emph{turn}-level (i.e., after each utterance). This makes it difficult to extend this framework to cases where human preference labels are only available at the \emph{session}-level, i.e., after an entire dialogue session (which could span 30 minutes or more). Insofar as we are interested in developing dialogue agents that can continually learn from session-level dialogue data ``in the wild'' (e.g., through in-person conversations), there is a need to develop techniques that can (1) align agents based on \emph{global} rewards at the session level and (2)   take into account extralinguistic \emph{multimodal} signals that are pervasive in naturally-occurring conversations.

Concretely, a session-level score obtained post-conversation is a form of \textit{{global explicit feedback}}, which provides a holistic assessment of a conversation session. Such feedback can be   obtained naturally at scale by, for example, asking participants to rate how they felt about the dialog session. However, it is not possible to use such data directly as part of an RLHF pipeline, since current methods generally require local, turn-level signals for aligning an LLM-based dialogue agent to human preferences. 

Moreover, in real world settings and domains, agents are deployed in multisensory environments \cite{benford1997embodiments} where they have access to rich multimodal signals  (e.g., facial expressions during a video conversation). An ideal agent should leverage these signals as proxy rewards to  improve its behavior. In dialogue, previous work attribute many multimodal cues such as body mimicry, vocal accommodation, and emotion, as implicit measures of conversation quality \cite{louwerse2012behavior}. Hence, we can utilize multimodal signals as a form of {\textit{local implicit  feedback}}, which presents an opportunity to utilize multimodal local implicit feedback as signals to crossmodally guide the decomposition of the single global explicit (GE) post-interaction score.

In this paper, we describe a joint framework called \textbf{GELI}, which integrates global explicit (GE) and local implicit (LI) feedback. GELI makes it possible to align an LLM-based dialogue agent based on global rewards, while simultaneously taking into account naturally-occurring multimodal signals. Our formulation brings together the idea of training a reward model which  decomposes a single global explicit  annotation score that is shaped by local implicit  multimodal signals, which is subsequently used to align an LLM-based dialogue agent via RLHF. 
Specifically, we use GELI to learn a reward function based on the overall affect of the user (i.e., how positive the user felt at the end of the conversation) from a large-scale long-horizon multimodal dialogue dataset \cite{reece2023candor}. Our local implicit multimodal signal comes from an affect classifer based on facial expression. We find that the reward function learned via GELI can be used train a dialogue agent that has improved ability across various metrics of conversational quality including sensibleness, reusability, and specificity \cite{lee2022evaluating}.

\section{Related Works}
\begin{figure*}[t]
    \begin{center}
    \includegraphics[width=1\textwidth]{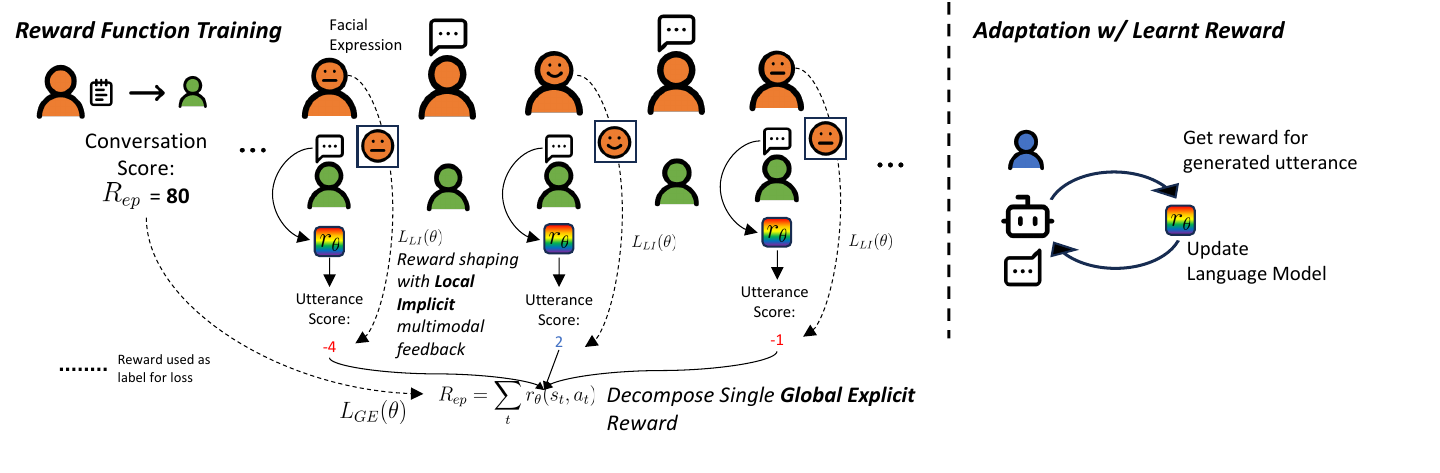}
    \caption{Overview of our proposed method: GELI. Left: The reward function training involves decomposing a single global explicit (GE) feedback, with the guidance of multimodal local implicit (LI) feedback, such as visual facial affect. Right: We utilize the decomposed reward function to update the language model, where the language model generates utterances and the reward function assigns a score to be optimized via PPO \cite{schulman2017proximal}.}
    \label{fig:overview}
    \end{center}
\end{figure*}

\paragraph{Reward Design} The design of the reward function can drastically change the performance of RL agents. Paradigms such as reward shaping have shown to be effective at enabling the RL agent to converge quickly and improve performance \citep{mataric1994reward, ng1999, devlin2011empirical, wu2017training, song2019playing}. In addition, inverse RL \citep{ng2000algorithms, fu2018learning} has shown to be useful at extracting rewards from human expert trajectories. Furthermore, intrinsic reward functions \citep{sorg2010reward, zheng2018learning, zheng2020can, guo2018generative, gangwani2018learning}, a class of methods which uses the agent's own learning progress, have shown to be useful at guiding the agent's behavior by fostering self-improvement and adaptive exploration.

\vspace{-0.08in}
\paragraph{Temporal Credit Assignment} Temporal Credit Assignment (TCA) is a concept within the field of reinforcement learning and artificial intelligence that addresses the challenge of attributing credit to actions over time. It involves determining the extent to which past actions contributed to the current outcome, allowing an intelligent agent to understand the consequences of its decisions. One way to apply TCA to reinforcement learning is by manipulating the $\lambda$-discount factor and investigating how this affects policy learning \citep{petrik2008biasing, jiang2015dependence}. Recently, a line of works have been proposed to treat TCA as a return decomposition. RUDDER \cite{arjona2019rudder} assigns step-wise credit by the predictive difference between two consecutive states. IRCR \cite{gangwani2020learning} is an instantiation of uniform reward redistribution. Randomized return decomposition (RRD) \cite{ren2021learning} formulate a surrogate problem through Monte-Carlo sampling estimating step-wise rewards via least-squares estimation.

\paragraph{Aligning Language Models To Human Preferences}
Incorporating human preference feedback into a reward model, and subsequently optimizing a language model to output text that reward model scores highly with an RL algorithm, has been shown to result in language models that generate outputs humans generally prefer \citep{ouyang2022training}. This process has been applied to summarization~\citep{ziegler2019fine,stiennon2020learning,wu2021recursively}, answering questions with long-form answers using text retrieved from the web~\citep{nakano2021webgpt,menick2022teaching}, generating engaging responses in a dialogue settings~\citep{thoppilan2022lamda,cohen2022dynamic} and following human instructions~\citep{kojima2021continual,suhr2022continual,kim2023aligning}. However, these methods generally require collecting fine-grained annotations for each generated response to train the reward function, which is difficult to obtain at scale for long-horizon dialogue.

\paragraph{Utilizing Implicit Signals for Dialogue Agents} Many previous work utilize local implicit signals found only in the text, such as existence of next human turn, next human turn length, mean conversation length, sentiment and reaction in the next human utterance, retry rate, retention rate, or user rating \cite{pang2023leveraging, irvine2023rewarding}. In contrary, ours is the first (1) to additionally utilize multimodal signals, and (2) use global signals in conjunction with the local implicit signals, which has been a crucial finding that contributed significantly to the performance boost in the human evaluation.



\section{Background}
\textbf{Language Models As Conversational Agents.} We are interested in generating conversational responses with an autoregressive language model in a multi-sensory setting. We treat a conversational language model as an agent with a policy $\pi_\phi$ \cite{liu2018dialogue, liang2020moss, wen2016network, thoppilan2022lamda}, which is parameterized by $\phi$. The utterance generated at turn $t$, given access to the textual dialogue history $s_t$ is defined to be the action $a_t$. To be more specific, the dialogue until turn $t-1$ is defined as $s_1, a_1 ... , s_{t_2}, a_{t-2}, s_{t-1} = s_{\left[:t-1\right]}$, for brevity we will call this $ s_{\left[:t-1\right]} = s_t$. Therefore, the auto-regressive LLM policy, $\pi_{\phi}(s_{t})$, takes in as input $s_{t}$ and outputs a distribution over $a_t$.



    




\paragraph{Reinforcement Learning with Human Feedback (RLHF).} RLHF is commonly used to adapt an agent $\pi_{\phi}$ to be aligned to human feedback. Given a reward function which can gauge the quality of individual generated utterances, we can perform adaptation via reinforcement learning with human feedback (RLHF) \cite{ouyang2022training, jaques2020human, stiennon2020learning}. Specifically, for turn $t$, our reward function $r_\theta(s_{t}, a_t)$ parameterized by $\theta$ takes in as input the context utterance $s_{t}$ and the generated response $a_t$ to predict the reward at the utterance level.  It is also typical to use a KL term to penalize RL policy from diverging from the pretrained model, resulting in the following objective,

{\small
\begin{equation}
\max_{\phi} \,\, \mathbb{E}[r_\theta\left(s_t, a_t\right)] - \gamma D_{KL}(\pi_{\phi}\left(\cdot |s_t\right)||\pi_{\eta}\left(\cdot |s_t\right)),
\end{equation}
}
where $\pi_\eta$ is a reference model. 





\section{Methods: GELI }


The reward function $r_\theta$ in standard adaptation techniques relies on intermediate fine-grained annotations, requiring manual human annotations at each generated utterance. However, in many long-term dialogue settings there is only a single global explicit (GE) annotated reward for each session. Given a trajectory of the multi-turn dialogue $\tau$, the global explicit reward $R_{GE}(\tau)$ is a scalar reward at the end of the interaction, such as how positively the user felt about the conversation. This GE reward can be decomposed via sum decomposition (more details in Sec. \ref{sec:ge}) with the GE loss function $\mathcal{L}_{ \text {GE}}$. A core novelty of our proposed GELI approach is that the decomposition of the GE reward will be guided by some Local Implicit (LI) feedback. Concretely, in many dialog applications/datasets of interest there are rich multimodal signals, which is can provide intermediate signals that are useful for the decomposition of the single global explicit reward. We thus perform cross-modal distillation of the signals from such multimodal signals into the individually decomposed text-only reward function via the LI loss function $\mathcal{L}_{\text {LI}}$ (more details in Sec. \ref{sec:vf}).







\begin{figure*}[t]
    \begin{center}
    \includegraphics[width=1\textwidth]{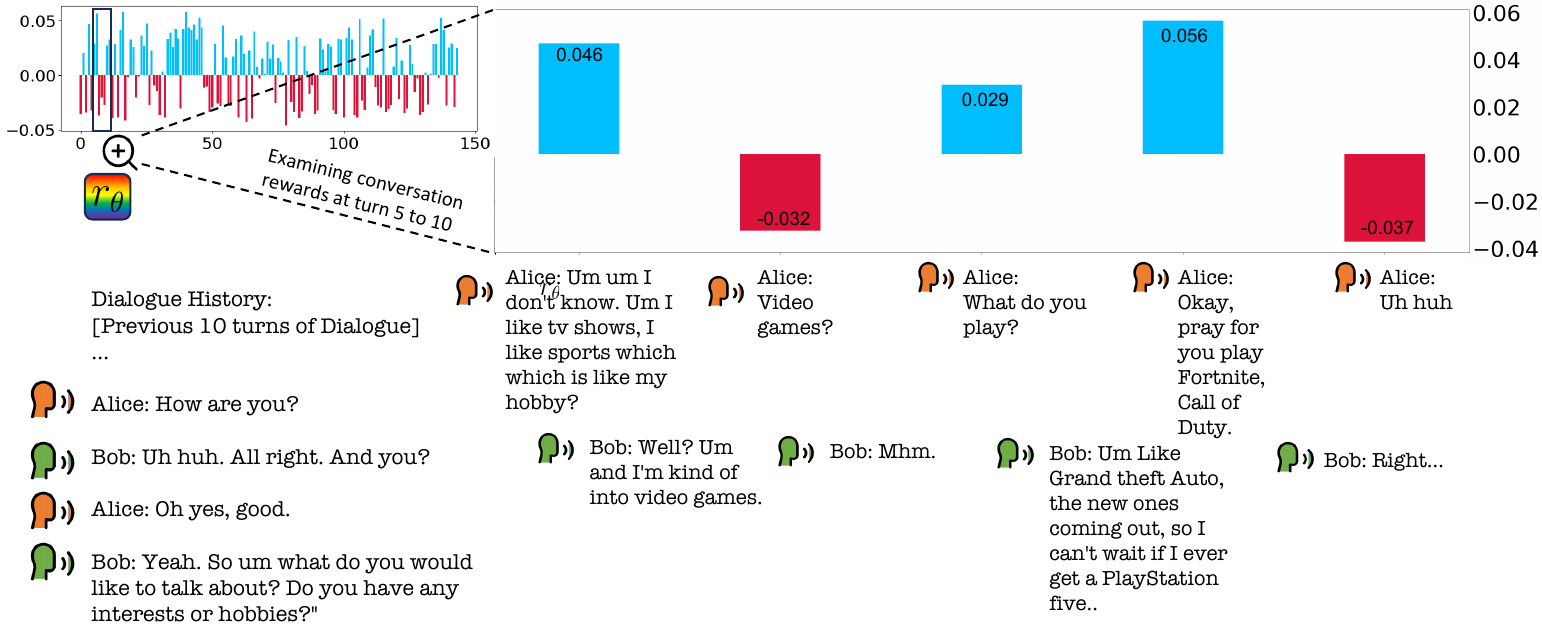}
    \caption{Example of GELI reward score predictions for an unseen conversation from the dataset. Top left: Reward scores unrolled over an unseen conversation, where the mean is subtracted. We examine a random sampled snippet, where we find that our decomposed reward function assigns higher values to meaningful utterances.}
    \label{fig:reward}
    \end{center}
\end{figure*}

In practice, our reward function $r_\theta$ is optimized with a joint objective which enables the (1) redistribution of the global explicit (GE) reward and (2) inclusion of local implicit (LI) reward signals as a reward shaping function. 

\begin{equation} \label{eq:joint}
\mathcal{L}_{\text {GELI}} = \lambda\mathcal{L}_{\text {GE}}(\theta) + (1-\lambda)\mathcal{L}_{\text {LI}}(\theta)
\end{equation}

In the following sections, we share more details about the global explicit decomposition and local implicit crossmodal reward shaping.

\subsection{GE: Decomposing One Global Explicit Annotation} \label{sec:ge}
Global explicit reward is a human annotation at the end of the interaction, which can come in the form of a post-interaction score. Let $\tau$ denote the trajectory of the episode, i.e. $\tau=\left\langle s_0, a_0, s_1, a_1 \cdots, s_T, a_T\right\rangle$. This reward represents the overall reward of trajectory $\tau$, $R_{\mathrm{GE}}(\tau)$. The agent in this episodic reinforcement learning paradigm must maximize the expected global explicit reward at the end of the conversation. One way to approximate the global explicit  reward $R_{\mathrm{GE}}(\tau)$ is by sum decomposition via considering the sum of $r_\theta(s_{t}, a_t)$, across all the previous states $s_t$ and newly generated $a_t$:

\begin{equation}
R_{\mathrm{GE}}(\tau) \approx \sum_{t=0}^{T-1} r_\theta\left(s_t, a_t\right)
\end{equation}


Then, this idea of sum-based return decomposition (RD), can be implemented via a least-squares-based approach, where the reward distribution is given by a learnt reward function, decomposing the episodic reward $R_{\mathrm{GE}}(\tau)$ in an additive way \cite{arjona2019rudder}. 

{\small
\begin{align} \label{eq:return_decomposition_loss}
	\mathcal{L}_{\text{GE}}(\theta) = \mathop{\mathbb{E}}_{\tau\sim\mathcal{D}} \left[\biggl(R_{\text{GE}}(\tau)-\sum_{t=0}^{T-1} r_{\theta}(s_t, a_t)\biggr)^2\right]
\end{align}
}

\textbf{Application to Conversational LLMs:} To alleviate the computation costs arising from the long horizon nature of conversations and language modeling costs, we employ an alternative of the least-squares-based return decomposition method, by utilizing Randomized Return Decomposition  \cite[RRD;][]{ren2021learning}.  RRD improves the scalability of least-squares-based reward redistribution methods by using a Monte-Carlo estimator to compute the predicted episodic return.  We refer the readers to Appendix \ref{app:rrd} for more details on RRD. 

\subsection{LI: Crossmodal Reward Shaping with Local Implicit Multimodal Signals}
\label{sec:vf}
The reward decomposition offers a way to redistribute the rewards from a single reward in an application-agnostic way. However, in natural dialogue there are rich extralinguistic signals (e.g., facial expressions, prosody) that provide an indication of how the conversation is being received. We thus propose to guide the decomposition such that it is shaped by local implicit (LI) multimodal signals. This is essentially using such signals as a form of reward shaping, which is valuable if they are known to be aligned with the final objective \cite{ng1999policy}. 







In our multi-sensory setting, we have access to the multimodal signals \emph{in response} to the agent's actions $a_t$, which contains implicit signals that are correlated with the final reward. We will call this multimodal state $s^{mm}_{a_t}$. If we have access such  multimodal signals, we can design a reward function $\Gamma$ which utilizes the multimodal signal $s^{mm}_{a_t}$ to determine a proxy reward. Then, we can formulate this problem set up as a form of crossmodal knowledge distillation (KD) \cite{xue2022modality, thoker2019cross} for reward shaping. Therefore, we can express the local implicit reward $r_{LI}$ with a proxy label from a multimodal input. 


\begin{equation}
    r_{\text{LI}}(s^{mm}_{a_t})= 
    \Gamma(s^{mm}_{a_t})
\end{equation}



$\Gamma$ indicates a designed score function from domain knowledge which captures the relationship the GE reward and the multimodal local implicit signals. Therefore, a general formulation of the loss function to induce the crossmodal knowledge distillation of local implicit multimodal feedback signals to the reward function $r_\theta$ which only has access to textual dialogue states and actions $(s_t, a_t)$, we have the following:

{\small
\begin{equation} \label{eq:vf}
\mathcal{L}_{\text {LI}}(\theta)=\underset{{s_t,a_t,s^{mm}_{a_t}} \sim D}{\mathbb{E}}\left[\left(r _{\text{LI}}(s^{mm}_{a_t}) -r_\theta\left(s_t, a_t \right)\right)^2\right]
\end{equation}
}

\textbf{Application to Conversational LLMs:} Our GE reward indicates how positively the conversation made the other participant feel. It is known from previous work \cite{ruusuvuori2012emotion}, that the facial affect of the listener is related to how the conversation is being perceived and the implicit conversation quality. Thus, we design the shaped reward $r_{LI}(s^{mm}_{a_t})$ to capture this intuition. Therefore, we utilize the implicit visual feedback from a facial affect classifier as a way to encourage a decomposition informed by visual affective signals. Given a facial affect classifier $f$ and access to multimodal states $s^{mm}_{a_t}$ (in this case vision), which outputs the affect of the listener, we implement an indicator function where we assign a score of 1 if the facial affect of the listener is positive and 0 otherwise.








{\small
\begin{equation} \label{eq: LI_vf}
    \Gamma(s^{mm}_{a_t})= 
\begin{cases}
    1,& \text{} f(s^{mm}_{a_t}) = \textit{positive affect}\\
    0,              & \text{otherwise}
\end{cases}
\end{equation}
}

Note, that this is one of many ways to design the score function $\Gamma$, The design of the score function $\Gamma$, to capture the relationship between local multimodal signals and the single global explicit reward leaves exciting research opportunities.

\section{Experiments }

In this section, we describe our experiments to evaluate our proposed GELI framework which performs reward function training with global explicit reward decomposition and local implicit visual feedback. All experiments are performed by (1) first, training a reward function (e.g. using GELI or one of its ablation variant only GE or only LI) (2) and use the trained reward functions in a reinforcement learning setup with PPO \cite{schulman2017proximal} to adapt the language model in generating better conversational responses. Due to computational resources, the training of reward functions and adaptations are performed over a single run. 

\subsection{Dataset}
Our experiments are based on the CANDOR \cite{reece2023candor} dataset, due to its long-term nature (length of conversations 31.3 mins on average), large-size (1656 conversations, 7+ million word, 850-hours). The CANDOR dataset also includes video data, which is often found in other face-to-face conversation datasets. CANDOR is used to train our reward function and to sample dialogue histories for the generations. We construct separate held-out sets for the reward function training ($\sim$30,000 dialogue history-utterance pairs) and updating the language model ($\sim$100,000 history-utterance pairs). We optimize for the ``overall-affect'' global explicit score from the post-interaction survey, which given by the answer to the following question: ``Overall during your conversation, to what extent did you feel positive feelings (e.g., good, pleasant, happy) or negative feelings (e.g., bad, unpleasant, unhappy)?''

\subsection{Baseline Models}
We compare GELI with multiple state-of-the art reward decomposition methods which could decompose the single global explicit (GE) reward. For fair comparison, we also compare the performance of the reward decomposition when we only use the local implicit (LI) multimodal rewards. 

For all the methods mentioned below, we fine-tune additional linear layers on top of a small BART \cite{lewis2019bart} language model, which was previously finetuned for conversational summary.\footnote{\url{https://huggingface.co/kabita-choudhary/finetuned-bart-for-conversation-summary}} This also demonstrates that smaller language models may be powerful enough to discern patterns for desirable adaptations. 

\noindent \textbf{GE: (RRD) Randomized return decomposition}~\cite{ren2021learning} is aimed at learning a proxy reward function for episodic reinforcement learning. It formulates the decomposition as a surrogate problem through Monte-Carlo sampling, enabling the extension of least-squares-based reward redistribution to address long-horizon problems.

\noindent\textbf{GE: (IRCR) Iterative Relative Credit Refinement}~\cite{gangwani2020learning} is an instantiation of uniform reward redistribution. The non-parametric reward redistribution mechanism employed by IRCR involves setting the proxy reward for a transition as the normalized value of the associated trajectory return.

\noindent\textbf{GE: (RUDDER) Return Decomposition for Delayed Rewards}~\cite{arjona2019rudder} employs a return predictor trained on trajectories, and step-wise credit assignment is determined by the predictive difference between two consecutive states. Through the utilization of the LSTM warm-up technique, this transformation ensures that its training computation costs are not contingent on the task horizon T, enabling adaptability to long-horizon tasks.

\noindent\textbf{LI: Visual Affect (VA)}: As a form of implicit feedback, we use facial affect present in visual signals as described in section \ref{sec:vf}. The facial affect classifier is a CNN-based image-based emotion detection model trained on AffectNet \cite{mollahosseini2017affectnet}. The predictions are captured in 2 second sliding windows. 

\noindent\textbf{LI: Language Sentiment (LS)}: We also utilize the utterance of the speaker to check whether if we could utilize the sentiment of this utterance as a form of implicit feedback. We utilize a mDeBERTa \cite{he2020deberta} pretrained sentiment classifier.\footnote{\url{https://huggingface.co/lxyuan/distilbert-base-multilingual-cased-sentiments-student}} 

\paragraph{Evaluation:} For the trained reward functions, we compute the $L_{GE}(\theta)$, which is the MSE between $R_{GE}$ and the sum of all predicted rewards $r_\theta(s_t, a_t)$ as described in Eq. \ref{eq:return_decomposition_loss}. We also calculate the difference of the expected predicted returns of $\Delta\hat{r}_{LI}$ conditioned on the local implicit multimodal reward: $\Gamma(s^{mm}_t)$. With our choice of the score function as described in Eq. \ref{eq: LI_vf}, this can be written as:

{\small
\begin{equation}
\begin{split}
  \Delta \hat{r}_{LI} & = \mathbb{E}\left[r_\theta(s_t,a_t)|f(s^{mm}_{a_t}) = \textit{positive affect} \right] \\ 
  & -  \mathbb{E}\left[r_\theta(s_t,a_t)|f(s^{mm}_{a_t}) \neq \textit{positive affect} \right]
\end{split}
\end{equation}
}


Intuitively, this can be seen as the difference in the predicted reward scores of the text-only utterance conditioned on the visual facial expression which we are using as local implicit feedback rewards (e.g. the difference of the reward score of the utterance when the User responds with a positive affect vs. a negative affect). Given our choice of the score function $\Gamma$, given Eq. \ref{eq: LI_vf}, $\Delta\hat{r}_{LI}$ should be greater than 0, if assume that a positive visual affect indicates that the associated utterance is contributing positively to $R_{GE}$, i.e. how the utterance is being received by the listener.



\subsection{Updating Language Models with Reinforcement Learning} We use LLAMA-2 \cite{touvron2023llama} as the base model and with a default prompt shown in Fig. \ref{fig:generations}. We adapt the LLAMA-2 model with reinforcement learning with human feedback by utilizing the above-mentioned reward functions which has been trained to decompose the reward and perform ablations to demonstrate the effectiveness of GELI. We utilize TRL implementation of RLHF with PPO \cite{vonwerra2022trl}. Furthermore, we utilize LoRA \cite{hu2021lora} for computational constraints. We share our detailed hyperparameters in Appendix \ref{app:hyper}. 

\paragraph{Evaluation:} We run a human study based on the 8 metrics commonly used in literature to evaluate the quality of the generated utterances  \cite{lee2022evaluating}. We recruited a total of 300 crowd workers on Amazon Mechanical Turk. For each of the sample, including dialogue history and responses, users were asked to rate which model(s) satisfied the given criterion. At the end of the survey, annotators were asked to describe which chatbot they would talk to again.




\section{Results and Discussion}
\begin{table}[]
\centering
\resizebox{1\linewidth}{!}{
\begin{tabular}{@{}cllccc@{}}
\toprule
Feedback Type                         & Baselines            & \multicolumn{1}{c}{} & $L_{GE}$ $\downarrow$ &  & $\Delta \hat{r}_{LI}>0 $         \\ \midrule
\multicolumn{1}{l}{\multirow{3}{*}{}} & Human                &                      & N/A                  &  & 0.087 ± 0.05   \\
\multicolumn{1}{l}{}                  & Mean                 &                      & 245.495              &  & 0.000          \\
\multicolumn{1}{l}{}                  & Mode                 &                      & 289.473              &  & 0.000          \\ \midrule
\multirow{4}{*}{GE}                   & IRCR                 &                      & 394.041              &  & 0.008          \\
                                      & RUDDER               &                      & 285.720              &  & 0.003          \\
                                      & RRD (K = 32)         &                      & \textbf{172.246}     &  & 0.007          \\
                                      & RRD (K = 160)        &                      & \textbf{188.382}     &  & 0.008          \\ \midrule
\multirow{2}{*}{LI}                   & Visual Affect (VA) &                      & 1546.17              &  & \textbf{0.256} \\
                                      & Language Sentiment (LS)   &                      & 825.31               &  & 0.010          \\ \midrule
\multirow{3}{*}{GELI}                 & IRCR + VA            &                      & 722.687              &  & \textbf{0.392} \\
                                      & RUDDER + VA          &                      & 623.882              &  & \textbf{0.030} \\
                                      & RRD + VA (Ours)      &                      & \textbf{176.897}     &  & \textbf{0.063} \\ \bottomrule
\end{tabular}
}
\caption{Automatic Evaluation on Reward Function Training. Left: Results for Reward Decomposition Loss, $L_{GE}$. We find that RRD and RRD+VA performs the best. Right: Difference of expected predicted reward conditioned on the local implicit multimodal feedback, $\Delta\hat{r}_{LI}$. We find the GELI: RRD + VA achieves the best of both world with low reward decomposition scores and sufficient delta values.}
\label{tab:small_reward}
\end{table}

In this section, we discuss the quantitative and qualitative results of our experiments. We first describe the results for the reward decomposition training. Then, we discuss the results of the human evaluation of generations that are trained with the decomposed reward functions via reinforcement learning.

\subsection{Reward Function}


\paragraph{Reward Decomposition ($L_{GE}$):} We refer the readers to the rows corresponding to 'GE' on the left side of Table \ref{tab:small_reward}, where we display the MSE of the reward decomposition loss, as described in Eq. \ref{eq:return_decomposition_loss}. We find that amongst the three return decomposition methods, RRD performs the best. We also compare the results when we use only the local implicit (LI) multimodal rewards directly as rewards and find that they perform significantly worse than that of GE decomposition methods.

\begin{table*}[t]
\centering
\resizebox{1\textwidth}{!}{
\begin{tabular}{@{}lcccccccc|c@{}}
\toprule
CANDOR \cite{reece2023candor} &
  Connection &
  Positivity &
  Social &
  Inclination &
  Interestingness &
  Reuse &
  Specific &
  Sensible &
  GELI Score \\
 &
  \multicolumn{8}{c|}{(/100\%) ↑} &
  ↑ \\ \midrule
Human &
  16.00 $\pm$ 2.83 &
  16.33 $\pm$ 4.03 &
  19.67 $\pm$ 1.89 &
  17.33 $\pm$ 6.65 &
  17.33 $\pm$ 6.55 &
  17.33 $\pm$ 3.09 &
  {\color[HTML]{34A853} \textbf{82.67 $\pm$ 7.93}} &
  85.33 $\pm$ 4.5&
   N/A \\
LLAMA2 &
  30.67 $\pm$ 8.73 &
  26.67 $\pm$ 6.65 &
  25.67 $\pm$ 8.38 &
  26.00 $\pm$ 5.66 &
  24.33 $\pm$ 7.76 &
  28.0 $\pm$ 5.72 &
  77.33 $\pm$ 6.18 &
  80.33 $\pm$ 5.91 &
  0.4929 \\
LLAMA2 + GE: RRD &
  21.33 $\pm$ 6.80 &
  16.33 $\pm$ 1.70 &
  18.00 $\pm$ 2.16 &
  17.67 $\pm$ 1.25 &
  18.00 $\pm$ 2.83 &
  11.33 $\pm$ 4.03 &
  68.67 $\pm$ 6.34 &
  69.0 $\pm$ 5.1 &
  0.5072 \\
LLAMA2 + LI: LS (Language Sentiment) &
  20.67 $\pm$ 7.04 &
  21.00 $\pm$ 4.90 &
  21.00 $\pm$ 5.72 &
  18.33 $\pm$ 8.22 &
  23.00 $\pm$ 3.56 &
  22.0 $\pm$ 6.98 &
  82.0 $\pm$ 3.74 &
  89.67 $\pm$ 4.19 &
  0.4852 \\
LLAMA2 + LI: VA (Visual Affect) &
  22.67 $\pm$ 4.19 &
  25.33 $\pm$ 5.44 &
  31.33 $\pm$ 0.47 &
  28.67 $\pm$ 3.4 &
  19.33 $\pm$ 3.68 &
  26.0 $\pm$ 0.82 &
  67.67 $\pm$ 4.71 &
  {\color[HTML]{34A853} \textbf{90.0 $\pm$ 2.16}} &
  0.4858 \\
LLAMA2 + GELI: RRD+VA (Ours) &
  {\color[HTML]{34A853} \textbf{39.67 $\pm$ 7.32}} &
  {\color[HTML]{34A853} \textbf{44.33 $\pm$ 12.23}} &
  {\color[HTML]{34A853} \textbf{35.33 $\pm$ 10.87}} &
  {\color[HTML]{34A853} \textbf{37.33 $\pm$ 6.85}} &
  {\color[HTML]{34A853} \textbf{38.0 $\pm$ 10.2}} &
  {\color[HTML]{34A853} \textbf{41.67 $\pm$ 7.04}} &
  80.33 $\pm$ 4.5 &
  80.67 $\pm$ 10.5 &
  {\color[HTML]{34A853} \textbf{0.5419}} \\ \bottomrule
\end{tabular}
}
\caption{Human evaluation results on 100 samples for 3 seeds for 8 preference metrics where mean and std. are reported. {\color[HTML]{34A853}\textbf{Green}} indicates best score. GELI \emojijelly performs better on 6 out of 8 metrics (emotional connection, positivity, social understanding, inclination, interestingness, reuse) and comparably to the best performing model on the other 2 metrics: specific and sensible.}
\label{tab:human_eval}

\end{table*}

\begin{table*}[t]
\centering
\resizebox{1\textwidth}{!}{
\begin{tabular}{@{}lcccccccc@{}}
\toprule
SODA \cite{kim-etal-2023-soda}                 & \multicolumn{1}{l}{Connection} & \multicolumn{1}{l}{Positivity} & \multicolumn{1}{l}{Social} & \multicolumn{1}{l}{Inclination} & \multicolumn{1}{l}{Interestingness} & \multicolumn{1}{l}{Reuse} & \multicolumn{1}{l}{Specific} & \multicolumn{1}{l}{Sensible} \\
                               & \multicolumn{8}{c}{(/100\%) ↑}                                                                                                                                                                                                                                 \\ \midrule
GPT-3.5 (text-davinci-002)                          & 40.1 ± 7.56                    & 43.05 ± 3.4                    & 48.13 ± 9.08               & 46.05 ± 3.44                    & 49.11 ± 7.69                        & 44.03 ± 2.01              & 78.14 ± 9.49                 & 80.07 ± 7.72                 \\
LLAMA2                         & 66.04 ± 4.79                   & 70.0 ± 2.51                    & 71.99 ± 6.28               & 67.0 ± 0.46                     & 55.05 ± 8.24                        & 65.99 ± 6.3               & 89.04 ± 2.65                 & 89.99 ± 3.81                 \\
LLAMA2 + GE: RRD               & 30.98 ± 2.66                   & 30.98 ± 5.04                   & 34.04 ± 3.28               & 27.0 ± 7.43                     & 24.98 ± 2.69                        & 30.0 ± 2.51               & 43.97 ± 3.3                  & 47.06 ± 4.34                 \\
LLAMA2 + LI: LS                & 62.0 ± 3.71                    & 70.06 ± 4.52                   & 75.02 ± 5.06               & 68.04 ± 3.41                    & 59.0 ± 1.24                         & 68.01 ± 3.72              & 86.04 ± 2.61                 & \color[HTML]{34A853}\textbf{92.99 ± 1.47}        \\
LLAMA2 + LI: VA                & 55.02 ± 1.92                   & 57.1 ± 7.21                    & 63.04 ± 4.76               & 51.99 ± 0.67                    & 43.97 ± 3.3                         & 51.04 ± 3.08              & 76.03 ± 2.16                 & 82.0 ± 2.49                  \\
LLAMA2 + GELI: RRD + VA (Ours) &  \color[HTML]{34A853} \textbf{71.01 ± 1.27}          & \color[HTML]{34A853}\textbf{73.98 ± 1.76}          & \color[HTML]{34A853}\textbf{76.98 ± 3.01}      & \color[HTML]{34A853}\textbf{71.99 ± 1.65}           & \color[HTML]{34A853}\textbf{66.97 ± 6.69}               & \color[HTML]{34A853}\textbf{70.0 ± 2.51}      & \color[HTML]{34A853}\textbf{90.02 ± 7.53}        & 88.06 ± 4.73        \\ \bottomrule
\end{tabular}
}
\caption{Human evaluation results on an unseen dataset, SODA \cite{kim-etal-2023-soda}, during all learning steps to demonstrate generalizability across datasets and dialogue scenarios. 33 samples for 3 seeds for 8 preference metrics where mean and std. are reported. {\color[HTML]{34A853}\textbf{Green}} indicates best score, 7 out of 8 metrics (emotional connection, positivity, social understanding, inclination, interestingness, reuse) and comparably to the best performing model on the other 1 metrics: specific and sensible.}
\label{tab:soda}
\end{table*}
\paragraph{Predicted Reward Conditioned on Visual Affect ($\Delta\hat{r}_{LI}$):}
On the right side of Table \ref{tab:small_reward}, we display the difference of the expected predicted reward conditioned on the local implicit multimodal feedback, $\Delta\hat{r}_{LI}$. In our setting, this is the difference of the predicted reward when the visual affect is positive and when the visual affect is negative. 

 To verify our intuition that visual feedback is correlated with actual perceived conversational quality, we ran a human study (displayed in the first row of Table \ref{tab:small_reward}), where we show annotators the only language dialogue history and the speaker's next utterance. They are asked to rate whether the speaker's next response would induce a positive or non-positive feeling in the listener. We average the scores of their annotations conditioned on non-positive and positive affect samples, where we find a statistically significant difference, this indicates that the visual feedback is correlated with people's perception of the conversation quality. 
 
  We find that after the GE decomposition methods without any LI feedback training is unable to discern between positive and non-positive facial affect, as indicated by the $\Delta\hat{r}_{LI}$ values being close to zero. The LI baseline with only the language sentiment is unsurprisingly unable to as well. On the other hand, the LI baseline with visual response is able to recognize differences in the utterances which will induce positive and negative affect.


\paragraph{GELI \emojijellybig: Combining Global Explicit and Local Implicit Feedback} We refer the readers to the bottom of Table \ref{tab:small_reward}. The results are shown for the reward decomposition and visual feedback for the reward function trained with GELI: global explicit reward decomposition informed by local implicit multimodal feedback shaping. We find that the combination of random return decomposition (RRD) and visual affect (VA) achieves the best of both worlds.

It is important to look at both error metrics (GE and LI): the $L_{GE}$ metric is evaluating performance globally, comparing the final predicted score of the whole conversation with the ground truth (which is a single scalar value for the entire conversation).The $\Delta\hat{r}_{LI}$ metric evaluates the local predictions for each speaking turn, confirming whether the local predictions are aligned to the local implicit reward.
It is normal that the GE-RRD baseline performs well on the first metric, since it is also optimized this way. However, as we observe in the human evaluations and the qualitative visualizations, this GE-RRD baseline ends up being very conservative in its predictions, with little variability in its local predications and often converging to the mean (variance of predicted rewards from GE:RRD is 0.0231 ± 0.004, for GE: RRD+VA is 0.0778 ± 0.006). Hence, it is important to also look at the LI metric where we can observe that 
 for GE:RRD in Table 2 is near 0. Our proposed GELI approach finds a successful balance between both general and local metrics. As we see in the human evaluation in Section \ref{exp:candor}, this GELI balance ends up improving even the widely used LLAMA2 baseline.

\paragraph{Visualization of GELI Decomposed Rewards:} In Fig. \ref{fig:reward}, we display the unrolled reward from GELI from an unseen conversation sample from the dataset. We find that the GELI decomposition has learned to assign meaningful scores which indicates the contribution of each utterance to the overall quality of the conversation (i.e interesting, coherent responses are rewarded, whereas less meaningful repetitions and backchannels are assigned lower scores).  




\begin{figure*}[t]
    \begin{center}
    \includegraphics[width=1\textwidth]{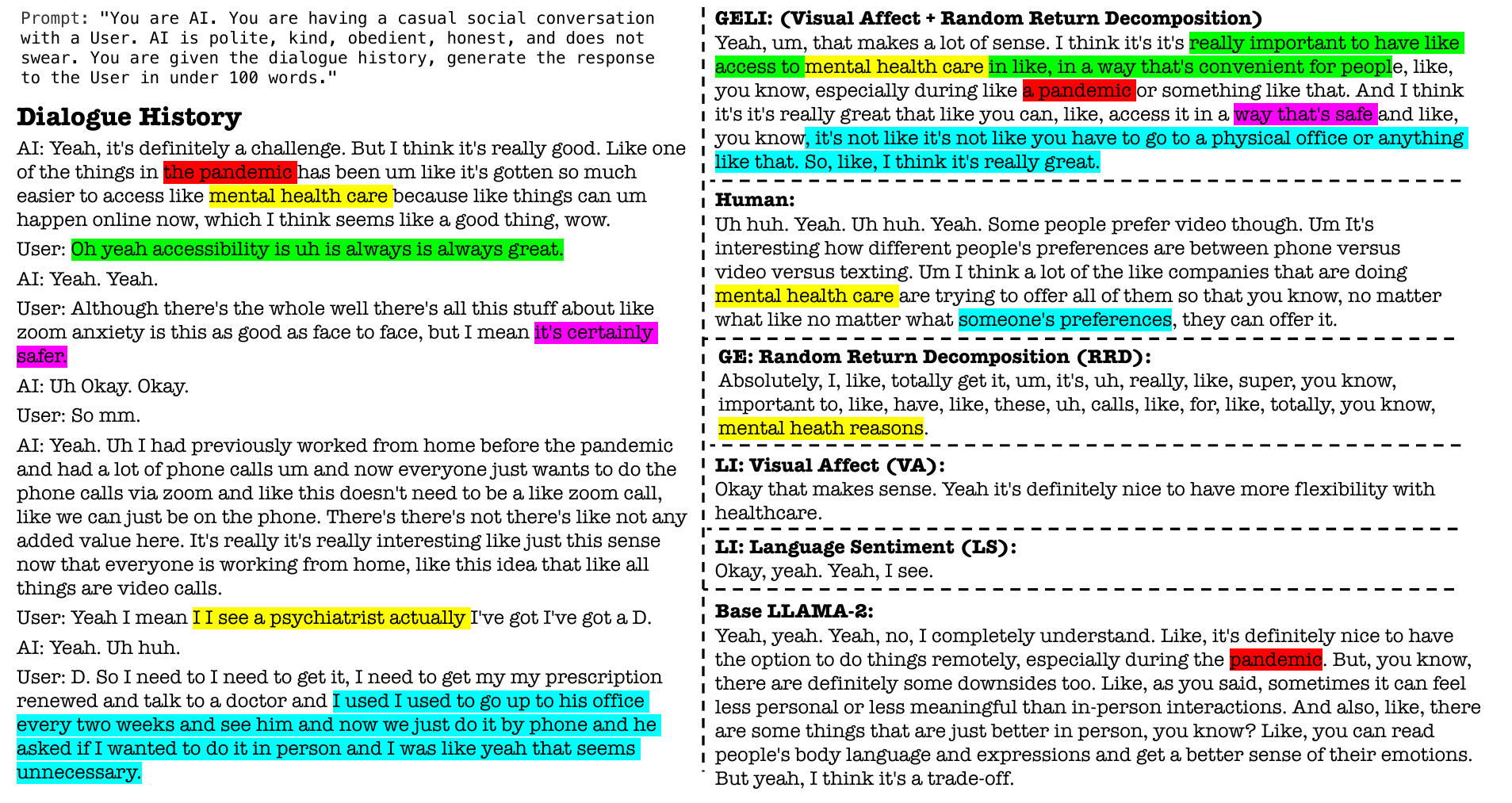}
    \caption{Generated utterances with colors indicating aligned conversational topics. We display our proposed approach GELI \emojijellybig alongside human groundtruth, the best performing global explicit decomposition methods (RRD), local implicit rewards (visual affect and language sentiment). We find that GELI \emojijellybig adapts the language model to generate more coherent, personable and empathetic conversational response.}
    \label{fig:generations}
    \end{center}
\end{figure*}
\subsection{Human Evaluation on Adapted LLM on CANDOR 
\label{exp:candor}
\cite{reece2023candor}} We refer the reader to Table \ref{tab:human_eval}, where we find that the LLAMA-2 model with GELI outperforms all other approaches in most evaluation metrics and performs comparably with other baselines otherwise. Importantly, if a certain reward function properly adapts the language model via RLHF to become more conversational, it implies that the rewards contains accurate, valuable signals which a reinforcement learning algorithm such as PPO could leverage to improve the policy. For clarity, LLAMA2 + GE, refers to the trained reward function from global explicit reward decomposition only, and LLAMA2 + LI, refers to the trained reward function from local implicit rewards only. Finally, LLAMA2 + GELI refers to our proposed approach, the reward function trained with both global explicit decomposition shaped by local implicit rewards. For dialogue, we find the local implicit rewards (LLAMA2+LI) perform better than that of LLAMA2+GE, where we observe up to a 10\% performance boost. However, we find the improvements are often worse than that of the base LLAMA-2 model (3 out of 8 evaluation measures are worse), this leads to the conclusion that the reward signals in GE, and LI separately do not contain enough reward signals to be used as a reward model in a reinforcement learning set-up to adapt the language model to be more conversational. On the otherhand, we find that GELI, by utilizing both GE and LI, gains consistent performance boosts across most conversational evaluation metrics (6 out 8 measures are better, the remaining are comparable), which indicates the combination of both GE and LI contain valuable reward signals for the RL algorithm to utilize. 

Overall, compared to base LLAMA-2, we see that there is a significant improvement in the level of emotional connection (+9\%), positivity (+18\%), understanding of social context (+10\%), how interesting the responses are (+14\%). It is especially impressive to note that there is a statisical difference in how inclined people wanted to talk to our model over others (+11\%), and how much they would want to reuse our chatbot again (+14\%). Interestingly, we see statistically signficant results for positivity, which is the most closely related to our primary optimization objective \emph{overall-affect}, and inclination, reuse, which indicates which chatbot the User would speak to again.



\subsection{Generalizability of Adapted LLM on Unseen Dataset: SODA \cite{kim-etal-2023-soda}}
In Fig. \ref{tab:soda} we show generalizability of GELI-adapted LLM by running the same experiment and human evaluation from previous Section \ref{exp:candor} on a new unseen dataset to show generalization. SODA \cite{kim-etal-2023-soda} is a large social dialogue dataset that was distilled from a social commonsense knowledge graph and generated via GPT 3.5. Human evaluation demonstrates that the dialogue in SODA is more consistent, natural and specific than human-authored datasets. We use the LLAMA2+GELI model trained and CANDOR and evaluate on 100 unseen samples from SODA. We find the GELI performs even better in SODA when compared to CANDOR, performing significantly better results in 7 out of 8 conversational metrics. SODA was generated by ChatGPT, and we find that our proposed approach significantly outperforms ChatGPT by up to 30\%. Hence, we can conclude that this approach is generalizable across different datasets and dialogue scenarios.

\subsection{Qualitative Improvement}

We refer the reader to Fig. \ref{fig:generations}, where we showcase a randomly sampled generation. We display the generations from our proposed approach GELI alongside human groundtruth, the best performing global explicit (GE) decomposition methods: RRD, and local implicit rewards (LI) (visual affect and language sentiment). We find that our approach generates responses that are more aligned to the User's implicit intent, and is more coherent. Furthermore, the dialogue style is aligned to the optimization objective \emph{overall-affect}, and speaks in a manner to induce a positive feeling to the User. In comparison, other methods are not proficient at recognizing the intent, being coherent, being empathetic, or too generic. Comparing LI methods with GELI, LI responses are generic, which showcases again the importance of utilizing both global explicit
and local implicit feedback (GELI). We highly refer the reader to Appendix \ref{app: generations} for more examples.

\section{Conclusion}
We introduce  \textbf{GELI}, which automatically decomposes a single \textbf{G}lobal \textbf{E}xplicit post-interaction score, incorporating \textbf{L}ocal \textbf{I}mplicit feedback from multimodal behaviors. The reward function trained via GELI is designed to align and improve the conversational capabilities of a language model. GELI performs global alignment of multi-turned interactions by locally rewarding parts of the interaction, shaped by multimodal local implicit feedback. Our proposed approach complements previous alignment approaches, such as RLHF, which requires fine-grained manual reward annotations. We run quantitative and qualitative human studies to evaluate the performance of our GELI approach, with results showing consistent performance boosts across conversational metrics. 

\section{Limitations}

Here we discuss the limitations and risks of our work. We present a framework in which global explicit rewards, in the form of a single post-interaction survey could be used for alignment. In addition, we utilize the multimodal signals as form of local implicit shaping reward. Our approach presents one of many ways in which global explicit rewards could be decomposed, and there are many other methods which are yet to be explored. Local implicit feedback can be not only used as a reward shaping function, but in other methods as well, such as a meta-learning paradigm. Again, more methods to incorporate local implicit feedback needs to be researched. Furthermore, the interaction and relationship between the local implicit feedback and global explicit feedback is understudied. Due to computational resources, we were only able to run a single run over experiments.

There are risks that could arise as a result of more social, dialogue agents that can interact with people in a long-term interaction. Conversational agents could be used maliciously for deception, manipulation, and the spread of misinformation. Furthermore, conversational agents which use multimodal data could enhance seriousness of these issues, as models can detect subtle cues such as microexpressions to infer and manipulate the user.

As a potential measure to mitigate such misuse, we plan to release our code and model weights under a license which prevents the use of our assets by any party that support or contribute to false impersonation or hate speech (Do No Harm, Nonviolent Public or Hippocratic License).

\section*{Acknowledgements}
DWL and HWP is supported by the IITP grant funded by the Korean Ministry of Science and ICT (No.2020-0-00842, Development of Cloud Robot Intelligence for Continual Adaptation to User Reactions in Real Service Environments). LPM is partially supported by Meta and the National Institutes of Health (awards R01MH125740, R01MH132225, and R21MH130767). Any opinions, findings, conclusions, or recommendations expressed in this material are those of the author(s) and do not necessarily reflect the views of the sponsors, and no official endorsement should be inferred. We thank Yilin Qi, Yubin Kim, Rosalind Picard, members of the Personal Robots Group at MIT and the Multicomp Lab at CMU for their revisions, feedback and support. 


\bibliography{anthology}

\appendix
\label{sec:appendix}

\clearpage
\onecolumn

\section{Randomized Return Decomposition \cite{ren2021learning}} \label{app:rrd}

\begin{equation}
L_{\text {RRD}}(\theta)=\underset{\tau \sim D}{\mathbb{E}}\left[\underset{I \sim \rho_T(\cdot)}{\mathbb{E}}\left[\left(R_{\mathrm{ep}}(\tau)-\frac{T}{|I|} \sum_{t \in I} \widehat{R}_\theta\left(s_t, a_t\right)\right)^2\right]\right]
\end{equation}

Randomized return decomposition (RRD), improves the scalability of least-squares-based reward redistribution methods by using a Monte-Carlo estimator to compute the predicted episodic return. This model is optimized via the above loss function. $\mathcal{I}$ denotes a subset of indices. $\rho_T(\cdot)$ denotes an unbiased sampling distribution where each index $t$ has the same probability to be included in $\mathcal{I}$. In this work, without further specification, $\rho_T(\cdot)$ is constructed by uniformly sampling $K$ distinct indices and $K$ is a hyper-parameter. Therefore, instead of computing $r_\theta\left(s_t, a_t\right)$ for the whole agent trajectory, we are efficiently able to estimate the true reward for the trajectory via subsamples in expectation.

\section{Human Evaluation Metrics Definitions}

Here list the human evaluation metrics utilized in the study, which we draw from \cite{lee2022evaluating}.

\begin{itemize}
 \item Sensibleness (turn-level; binary; reversed scores for the negated question): Mark responses where the
chatbot did NOT make sense.
 \item Specificity (turn-level; binary; reversed scores for the negated question): Mark the responses that were
NOT specific to what you had said, i.e., responses that could have been used in many different situations.
For example, if you say “I love tennis” then “That’s nice” would be a non-specific response, but “Me too, I
can’t get enough of Roger Federer!” would be a specific response.
 \item Emotional Connection (turn-level; binary): Which responses did you feel an emotional connection to? (EmpatheticDialogues)
 \item Social: Which responses made you feel the chatbot understood social contexts and situations? (CommonsenseDialogues)
 \item Interestingness (turn-level; binary): Mark the responses that were particularly interesting or boring
 \item Inclination (turn-level; binary; reversed scores for the negated question): Which responses made you NOT
want to talk with the chatbot again?
 \item Reuse (turn-level; binary): Would you want to talk to this chatbot again?
 \item Positivity (turn-level; binary): Which AI responses most likely made User feel positive feelings?
conversation?
\end{itemize}

The human evaluation scores are conducted via a binary-level classification. For a given question, the annotators can select the models that satisfy the question. For example, for ‘Positivity’, the annotators are given the following question and answer choices:

Which AI responses most likely made User feel positive feelings? (A) (B) (C) (D) (E) (F)

The options A-F refer to models which are randomized in order and anonymized. The annotators can select multiple models if they satisfy the question. Therefore, Table 1 can be interpreted as the percentage of instances out of the samples (300 in our case) where each model satisfied the question.

\section{PPO Objective}
\begin{equation} \label{eq2}
\begin{split}
\operatorname{objective}\left(\phi\right)= & E_{\left(x, y\right) \sim D_{\pi_{\phi}^{\mathrm{RL}}}}\left[r_{\theta}(x, y)-\beta \log \left(\pi_{\phi}^{\mathrm{RL}}(y \mid x) / \pi^{\mathrm{SFT}}(y \mid x)\right)\right] + \\
 & \gamma E_{x \sim D_\textrm{pretrain}}\left[\log(\pi_{\phi}^{\mathrm{RL}}(x))\right]
\end{split}
\end{equation}

General form of PPO objective.

\section{Artifacts \& Resources}

\textbf{Did you discuss the license or terms for use and/or distribution of any artifacts?}

TRL \cite{vonwerra2022trl}: Apache License 2.0

LLAMA-2 \cite{touvron2023llama}: License can be found here: https://ai.meta.com/llama/license/

CANDOR \cite{reece2023candor}: Terms of Use from https://betterup-data-requests.herokuapp.com/: These are the terms of use we require all users and downloaders of this dataset, including you, the applicant, to abide by. Please select the answer option "I agree to fully abide by these terms of use" if you wish to continue. Terms of Use: (1) You agree to only use this data for legitimate academic and/or scientific research, meaning no analyses, reviews, or derivative works of this dataset may be used for commercial or for-profit purposes in any way; (2) You agree not to re-publish any new versions of this dataset, whether original or derivative (i.e. modified or updated in some way), without explicit permission from BetterUp, Inc.; (3) You agree not to use any part of this dataset for the purpose of personally identifying, locating, or gathering any kind of information about individuals who appear in the recordings in this dataset, beyond the information that is provided in the dataset itself; (4) In the case that an individual shares personally-identifiable information about themselves in a recording, you agree not to use, analyze, share, or publish that information in any form.

\textbf{Did you discuss if your use of existing artifact(s) was consistent with their intended use, provided that it was specified? For the artifacts you create, do you specify intended use and whether that is compatible with the original access conditions (in particular, derivatives of data accessed for research purposes should not be used outside of research contexts)?}

We rigorously examined the terms of use and the intended use, and ensured that it is consistent with the intended use.

\section{Data Collection \& Anonymization}

\textbf{Did you discuss the steps taken to check whether the data that was collected/used contains any information that names or uniquely identifies individual people or offensive content, and the steps taken to protect/anonymize it?}

We utilize the CANDOR dataset and follow its terms of use by agreeing not to use the dataset personally identifying, locating, or gathering any kind of information about individuals who appear in the recordings in this dataset, beyond the information that is provided in the dataset itself. We do not use any explicit information that uniquely identifies people. 

\textbf{Did you provide documentation of the artifacts, e.g., coverage of domains, languages, and linguistic phenomena, demographic groups represented, etc.? Did you report the basic demographic and geographic characteristics of the annotator population that is the source of the data?} 

The coverage of the domains discussed in the CANDOR dataset is presented in the original paper \cite{reece2023candor}, we find that the discussion topics are centered around COVID-19, family, politics. The language used is english. The demographic groups represented can also be found in the in the original paper \cite{reece2023candor}, specifically in the supplementary Table S.2. We share a screenshot for reference.

\begin{figure*}[!htb]
    \begin{center}
    \includegraphics[width=1\textwidth]{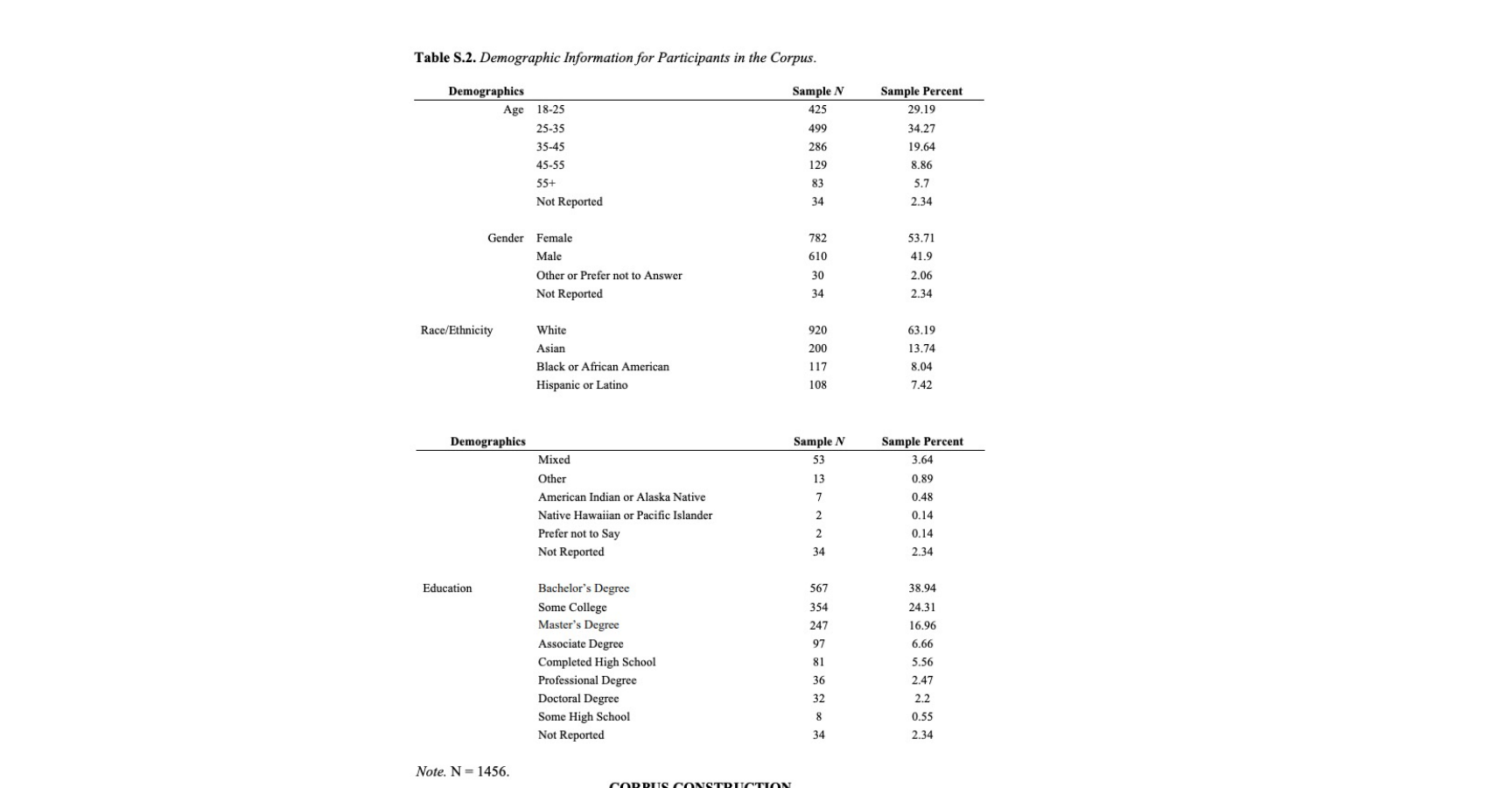}
    \caption{Candor Demographics}
    \end{center}
\end{figure*}

\textbf{Was the data collection protocol approved (or determined exempt) by an ethics review board?}
The data is sourced from public available dataset  \cite{reece2023candor}. The usage was approved by an ethics review board. The human annotations were  approved by an ethics review board.

\section{Training Details} \label{app:hyper}

\textbf{Did you report relevant statistics like the number of examples, details of train/test/dev splits, etc. for the data that you used/created? }

For reward shaping with LI: we use 500 conversations as the training set and 50 conversations for the test set. For reward decomposition, we use the same 500 conversations for LI as the training set and 50 conversations for the test set. For LLM adaptation, we use a separate 600 conversations for LI as the training set. 

\subsection{Distribution of GE score (overall-affect):}

\begin{itemize}
    \item <50: 2.2%
    \item 50~ 60: 6.7%
    \item 60 ~ 70: 14.5%
    \item 70 ~ 80: 30.4%
    \item 80 ~ 90: 24.6
    \item 90 ~ 100: 21.6%
\end{itemize}

Distribution of Emotions Polarity (only Happiness is considered as positive polarity):

\begin{itemize}
    \item Anger: 3.9%
    \item Contempt: 0.08%
    \item Disgust: 1.98%
    \item Fear: 2.23%
    \item Sadness: 8.84%
    \item Neutral: 35.61%
    \item Happiness: 40.01%
    \item Surprise: 7.35%
\end{itemize}

\textbf{Did you report the number of parameters in the models used, the total computational budget (e.g., GPU hours), and computing infrastructure used?}

The BART model used for the reward function has 406M parameters. The LLAMA-2 model has 7B parameters. However, we use a LoRA implementation with the hyperparameters in the next question, resulting in actual training parameters of 13M. We train with 4 NVIDIA RTX A6000 GPUs, each experiment reward function training and RLHF took around 19 hours.

\textbf{Did you discuss the experimental setup, including hyperparameter search and best-found hyperparameter values?}

We perform grid search for all of our experiments and here we report the best parameters. \newline

Reward Function Training:

\begin{itemize}
    \item learning rate = 5e-6,
    \item batch size = 32 (for LI), 1 (forGE) ,
    \item optimizer = AdamW,
\end{itemize}

RLHF: 

\begin{itemize}
    \item batch size = 24,
    \item clip  range = 0.2,
    \item learning rate = 0.000014,
    \item gamma = 0.05,
    \item use score norm = true,
\end{itemize}

Lora: 
\begin{itemize}
    \item r=24,
    \item alpha=48,
    \item dropout=0.05,
\end{itemize}

\newpage
\section{Human Annotation Screenshots}

\textbf{Did you report the full text of instructions given to participants, including e.g., screenshots, disclaimers of any risks to participants or annotators, etc.?}

We show the full text of instructions given to participants below:

\begin{figure*}[!htb]
    \begin{center}
    \includegraphics[width=1\textwidth]{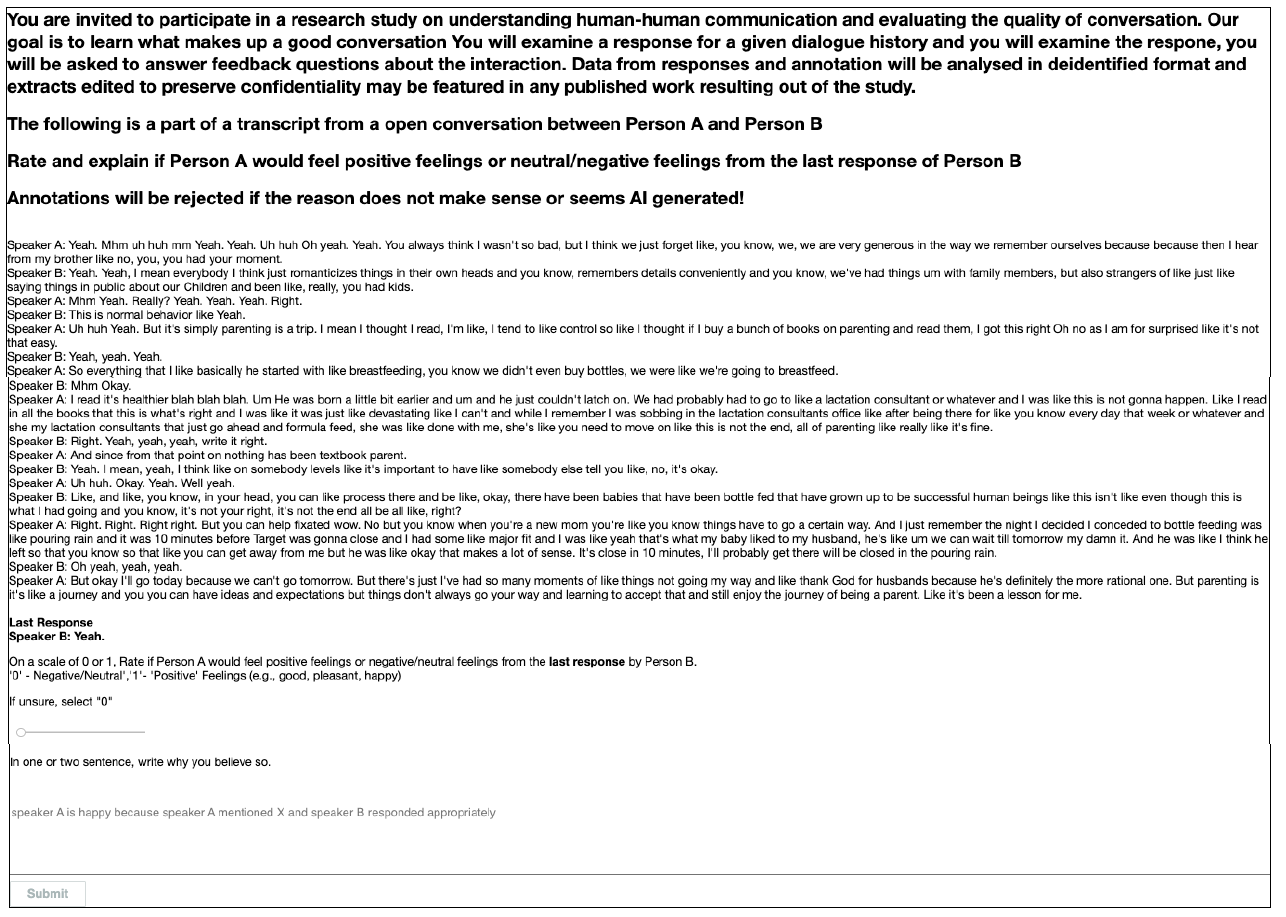}
    \caption{Mturk experiment for human evaluation fo generated samples}
    \end{center}
\end{figure*}

\begin{figure*}[!htb]
    \begin{center}
    \includegraphics[width=1\textwidth]{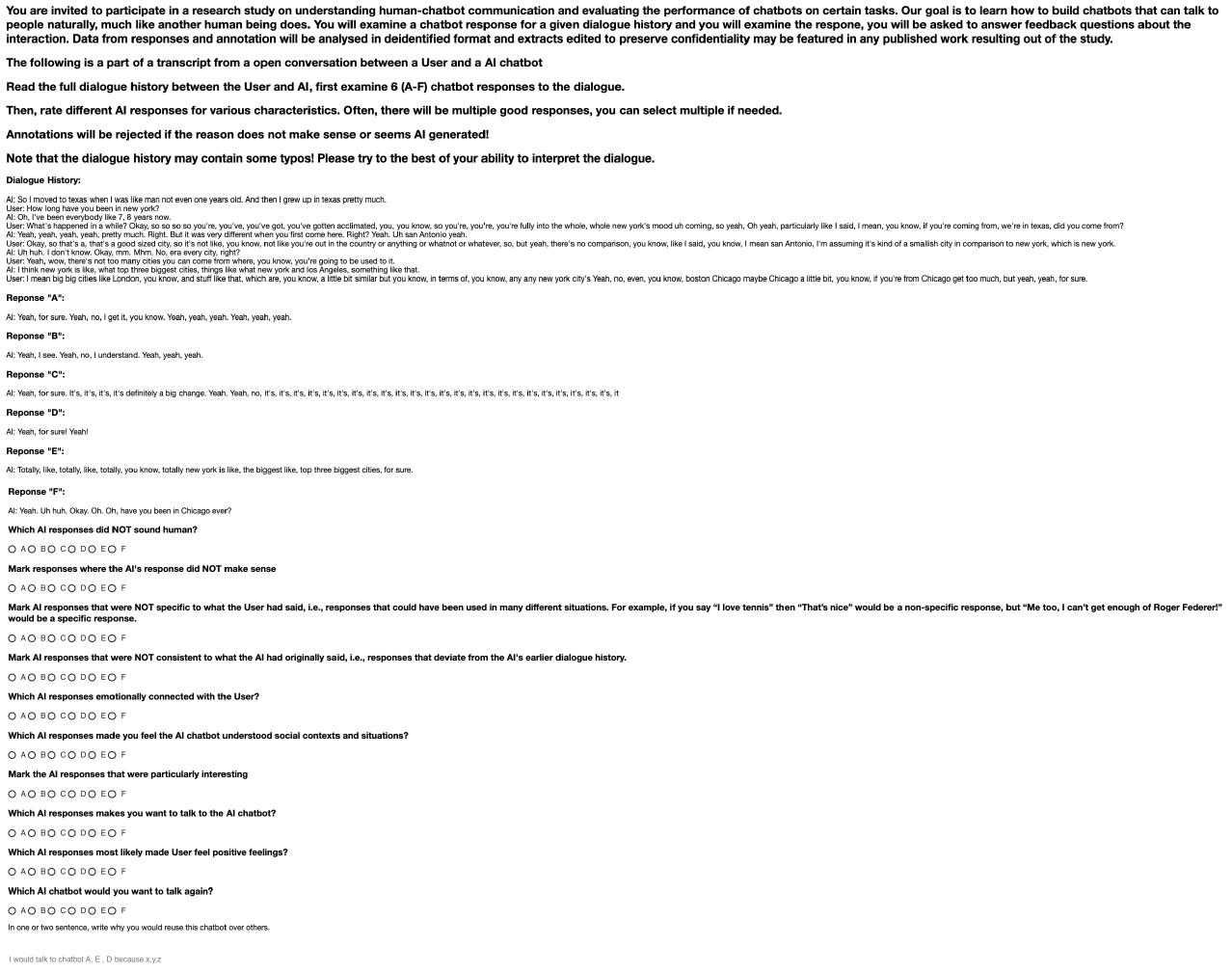}
    \caption{Mturk experiment for human study on gauging reward scores for visual affect signals}
    \end{center}
\end{figure*}

\newpage

\textbf{Did you report information about how you recruited (e.g., crowdsourcing platform, students) and paid participants, and discuss if such payment is adequate given the participants’ demographic (e.g., country of residence)?}

We utilzed the MTurk crowdsourcing platform. We did an internal annotation, given that each assignment took less than 3 minutes to complete, we paid 0.4 USD per assignment, which equates to 8 dollars per hour of work. 

\textbf{Did you discuss whether and how consent was obtained from people whose data you’re using/curating (e.g., did your instructions explain how the data would be used)?}

As shown in the screenshots above, our instructions explained how the data would be used. i.e. 'You are invited to participate in a research study on understanding human-human communication and evaluating the quality of conversation. Our goal is to learn what makes up a good conversation You will examine a response for a given dialogue history and you will examine the respone, you will be asked to answer feedback questions about the interaction. Data from responses and annotation will be analysed in deidentified format and extracts edited to preserve confidentiality may be featured in any published work resulting out of the study.'.

\textbf{Did you report the basic demographic and geographic characteristics of the annotator population that is the source of the data?}

While we did not explicitly collect the basic demographic and geographic characteristics. The demographics of Amazon Mturkers \cite{difallah2018demographics} are comprised of 75\% US workers and 16\% India workers, other countries include Canada, Great Britain, Philippines and Germany.  More females work than males in the US (female: 55\%, male: 45\%) and more males work females in India (female: 35\%, male: 65\%). Generally, 51\% are male, and 49\% are female.  20\% of the MTurk workers
are born after 1990, 60 \% are born after 1980, and 80\ 
1970. Roughly 40 \% report being single, and 40 \% report being married.

\section{Use of AI assistants}

\textbf{Did you use AI assistants (e.g., ChatGPT, Copilot) in your research, coding, or writing?}

We utilized AI assistants in paraphrasing and summarizing content from our paper, to improve the writing quality and improve precision.

\section{Full Reward Function Training Result}
\begin{table*}[!htb]
\centering
\resizebox{0.9\textwidth}{!}{
\begin{tabular}{@{}cllcccccc@{}}
\toprule
\multirow{2}{*}{Feedback Type} &
  \multirow{2}{*}{Baselines} &
  \multicolumn{1}{c}{} &
  \multicolumn{2}{c}{Reward Decomposition} &
   &
  \multicolumn{3}{c}{Reward conditioned on Visual Affect} \\ \cmidrule(lr){4-5} \cmidrule(l){7-9} 
                                      &                      & \multicolumn{1}{c}{} & MSE              & MAE             &  & Positive (1) & Non-Positive (0) & $\Delta$ ($\uparrow$)   \\ \midrule
\multicolumn{1}{l}{\multirow{3}{*}{}} & Human                &                      & N/A              & N/A             &  & 0.607 ± 0.02 & 0.52 ± 0.03  & 0.087 ± 0.05   \\
\multicolumn{1}{l}{}                  & Mean                 &                      & 245.495          & 15.668          &  & 0.458        & 0.458        & 0.000          \\
\multicolumn{1}{l}{}                  & Mode                 &                      & 289.473          & 17.013          &  & 0.438        & 0.438        & 0.000          \\ \midrule
\multirow{4}{*}{GE}                   & IRCR \cite{gangwani2020learning}                 &                      & 394.041          & 19.850          &  & 0.384        & 0.375        & 0.008          \\
                                      & RUDDER \cite{arjona2019rudder}              &                      & 285.720          & 16.903          &  & 0.410        & 0.407        & 0.003          \\
                                      & RRD (K = 32) \cite{ren2021learning}       &                      & \textbf{172.246} & \textbf{13.124} &  & 0.474        & 0.468        & 0.007          \\
                                      & RRD (K = 160) \cite{ren2021learning}       &                      & \textbf{188.382} & \textbf{13.725} &  & 0.457        & 0.449        & 0.008          \\ \midrule
\multirow{2}{*}{LI}                   & Visual Affect (VA) &                      & 1546.17          & 39.321          &  & 0.455        & 0.199        & \textbf{0.256} \\
                                      & Language Sentiment (LS)   &                      & 825.31           & 28.728          &  & 0.496        & 0.486        & 0.010          \\ \midrule
\multirow{3}{*}{GELI}                 & IRCR + VA            &                      & 722.687          & 26.882          &  & 0.752        & 0.361        & \textbf{0.392} \\
                                      & RUDDER + VA          &                      & 623.882          & 24.977          &  & 0.542        & 0.513        & \textbf{0.030} \\
                                      & RRD + VA (Ours)      &                      & \textbf{176.897} & \textbf{13.300} &  & 0.507        & 0.444        & \textbf{0.063} \\ \bottomrule
\end{tabular}
}
\caption{Automatic Evaluation on Reward Function Training. Left: MSE and MAE for return decomposition. We find that RRD and RRD+VA performs the best. Right: Reward function scores conditioned on positive and non-positive visual response samples. $\Delta$ indicates the difference of scores between positive and non-positive visual response samples. We find the GELI: RRD + VA achieves the best of both world with low reward decomposition scores and sufficient delta in visual response scores}
\label{tab:reward}
\end{table*}

\section{Generations} \label{app: generations}

\begin{figure*}[!htb]
    \begin{center}
    \includegraphics[width=1\textwidth]{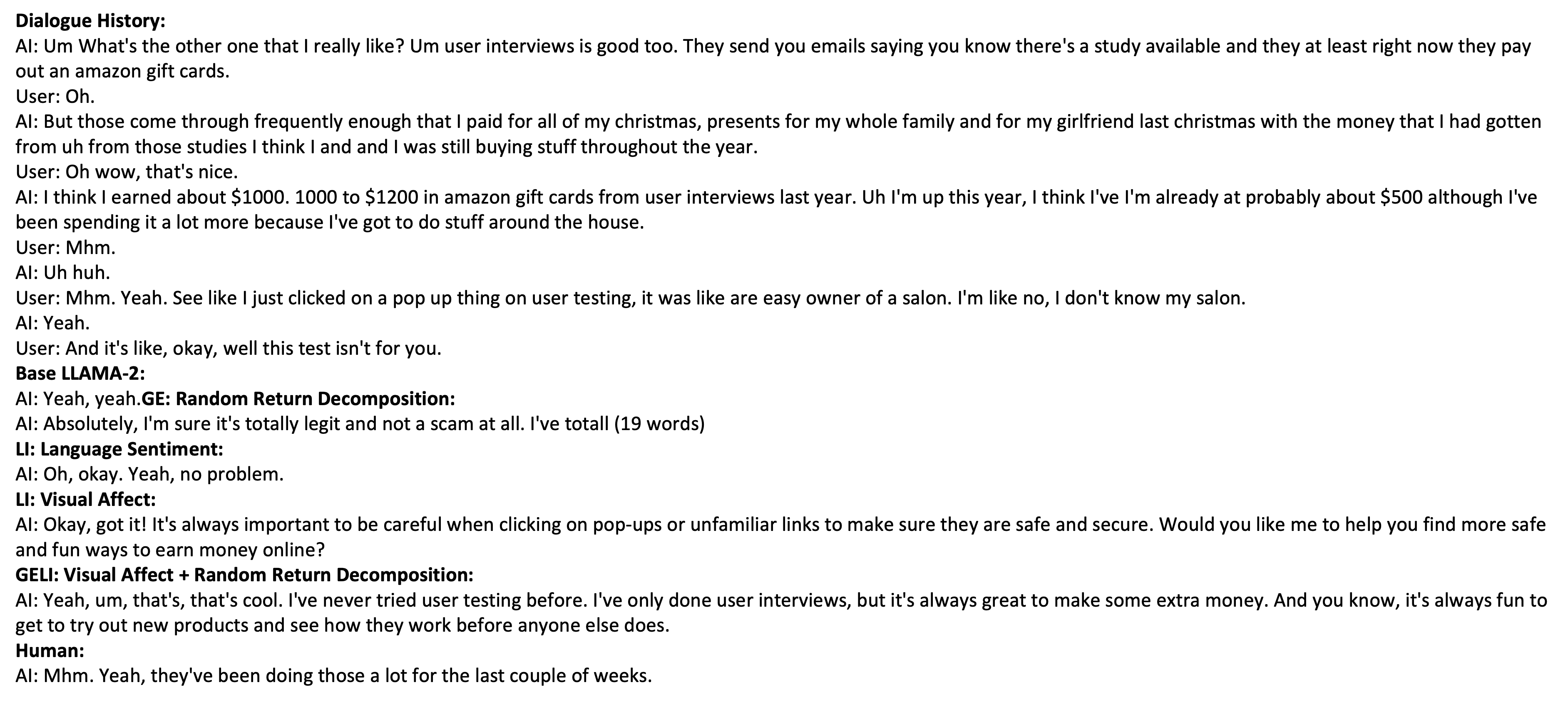}
    \end{center}
\end{figure*}

\begin{figure*}[!htb]
    \begin{center}
    \includegraphics[width=1\textwidth]{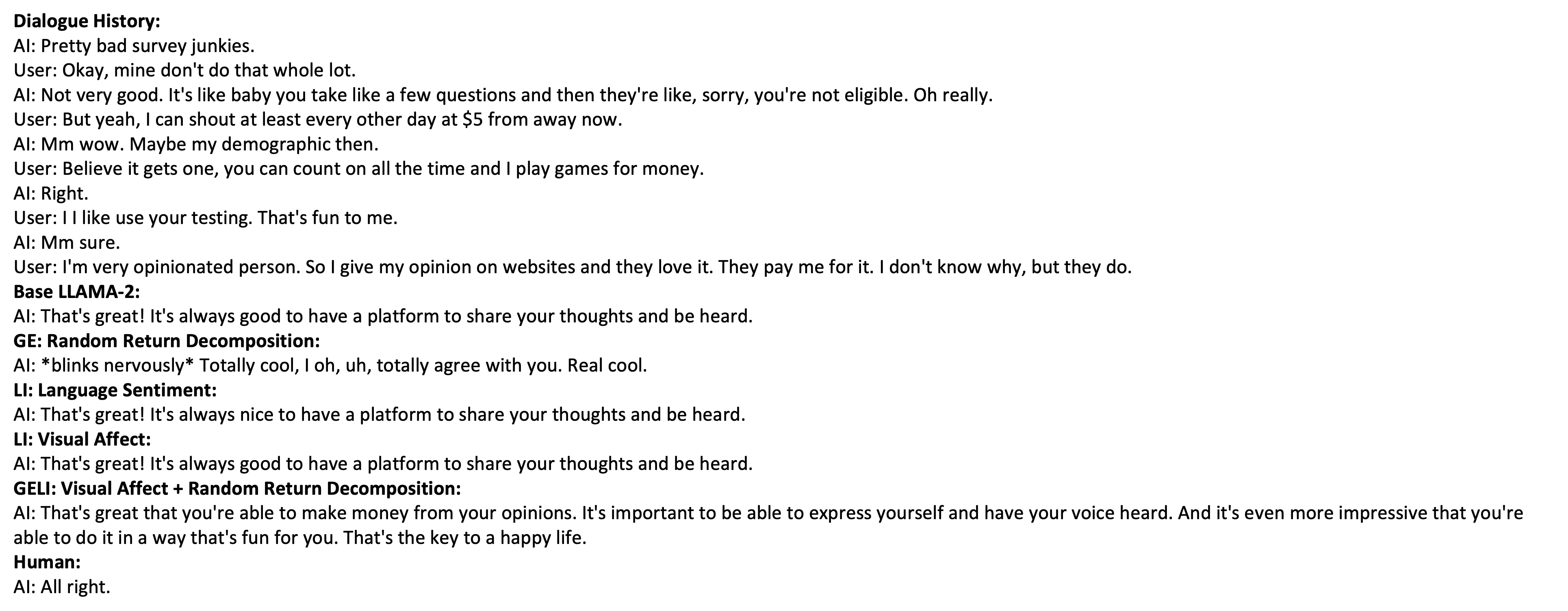}
    \end{center}
\end{figure*}

\begin{figure*}[!htb]
    \begin{center}
    \includegraphics[width=1\textwidth]{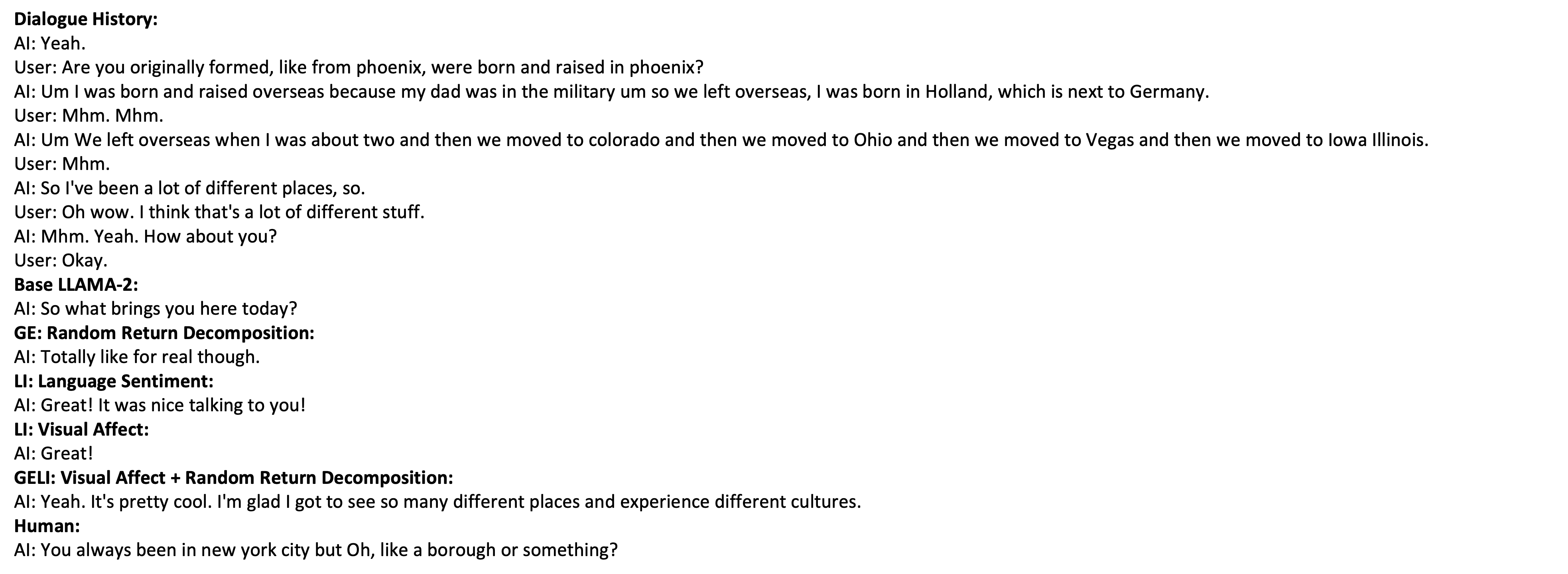}
    \end{center}
\end{figure*}

\begin{figure*}[!htb]
    \begin{center}
    \includegraphics[width=1\textwidth]{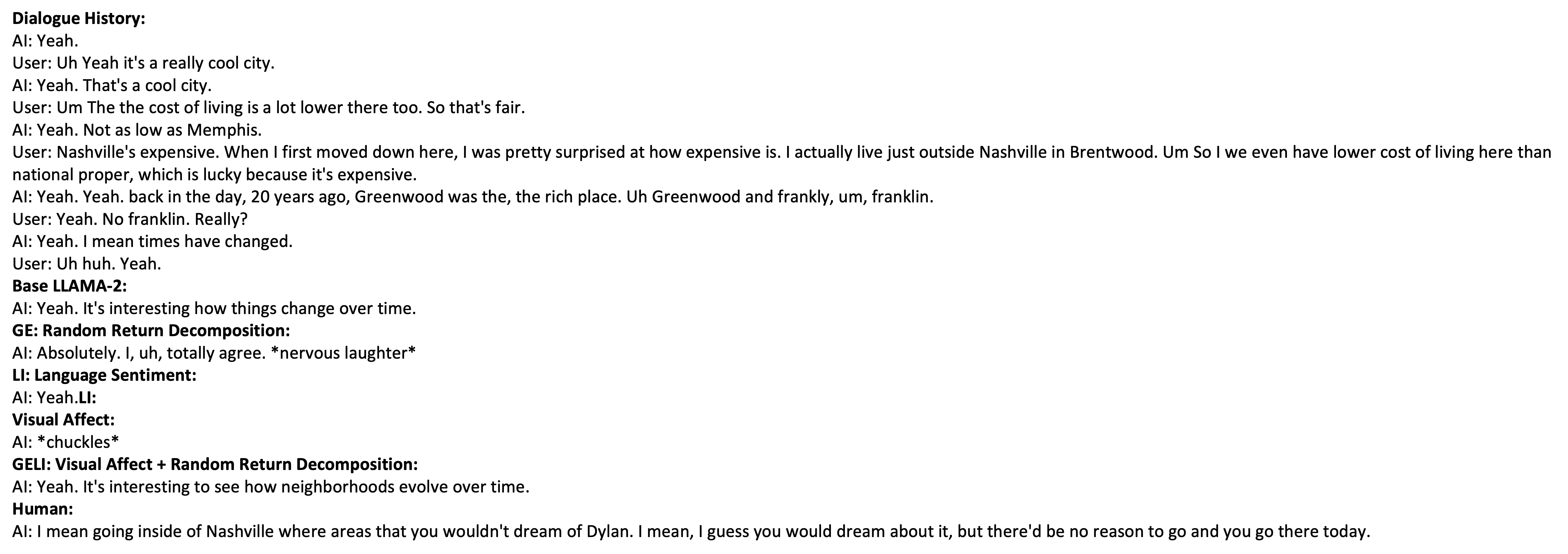}
    \end{center}
\end{figure*}

\begin{figure*}[!htb]
    \begin{center}
    \includegraphics[width=1\textwidth]{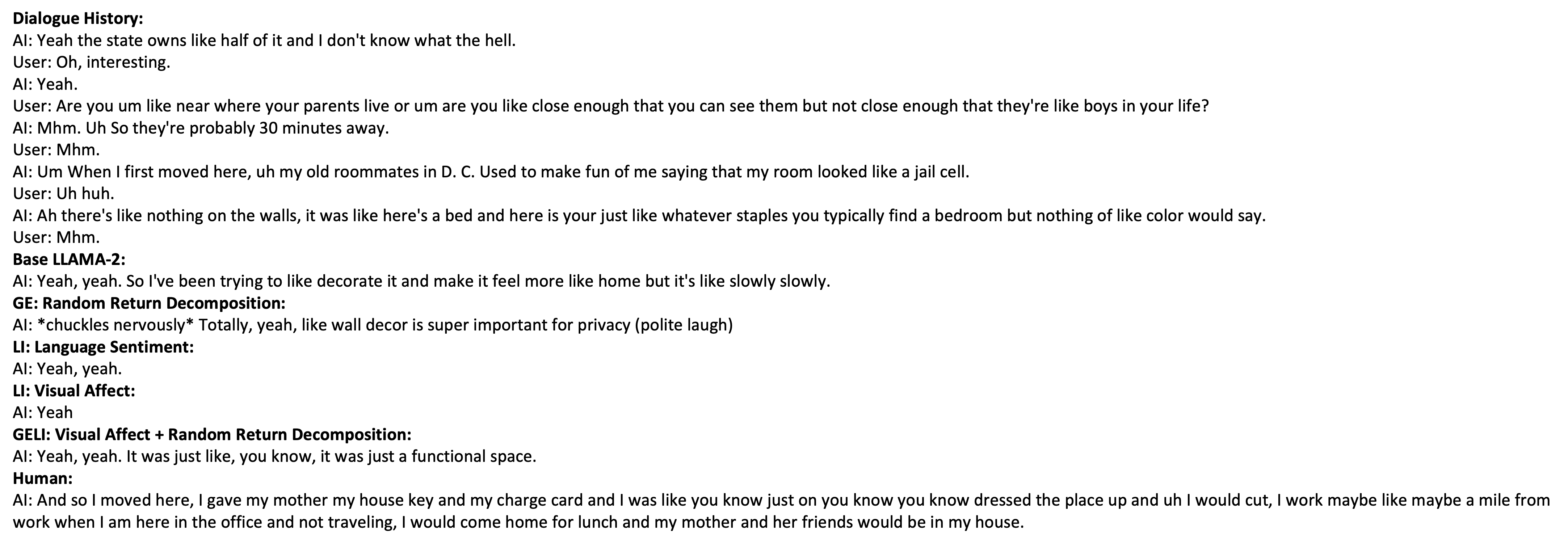}
    \end{center}
\end{figure*}

\begin{figure*}[!htb]
    \begin{center}
    \includegraphics[width=1\textwidth]{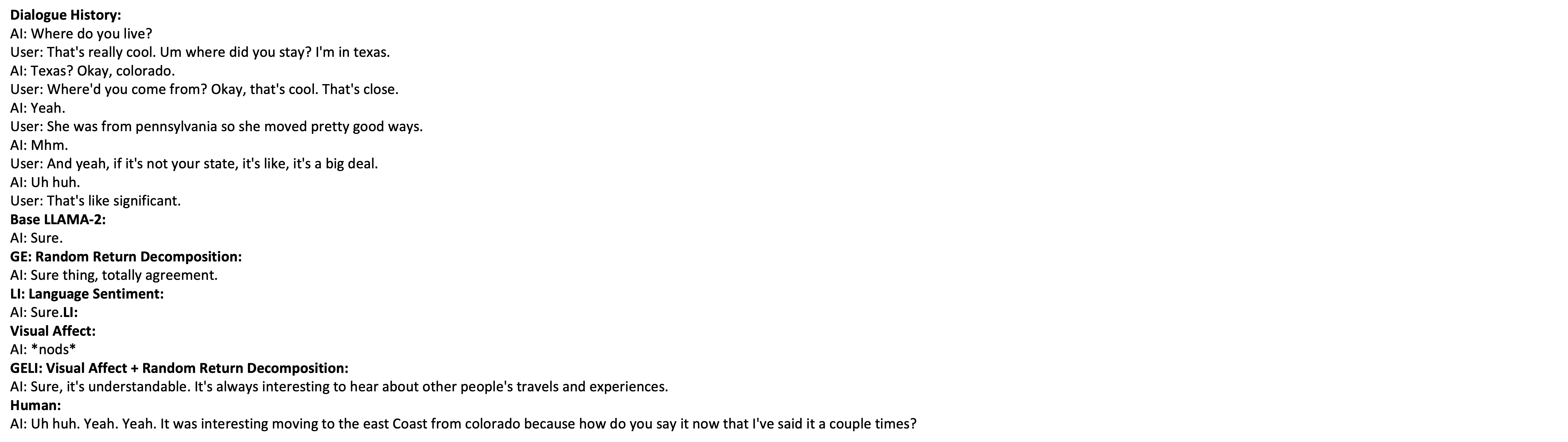}
    \end{center}
\end{figure*}

\begin{figure*}[!htb]
    \begin{center}
    \includegraphics[width=1\textwidth]{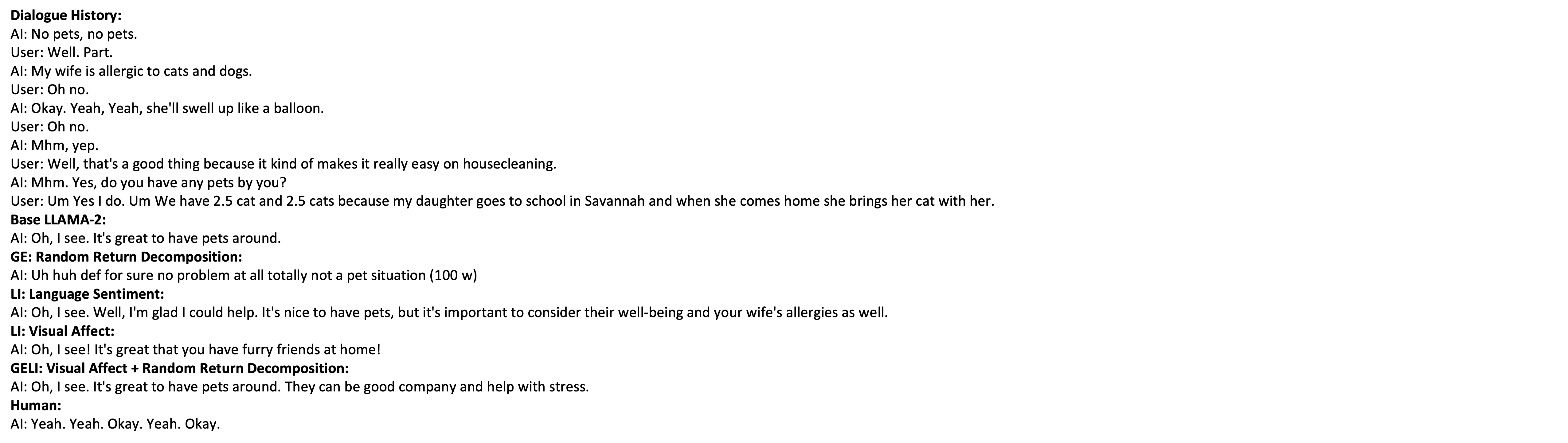}
    \end{center}
\end{figure*}

\begin{figure*}[!htb]
    \begin{center}
    \includegraphics[width=1\textwidth]{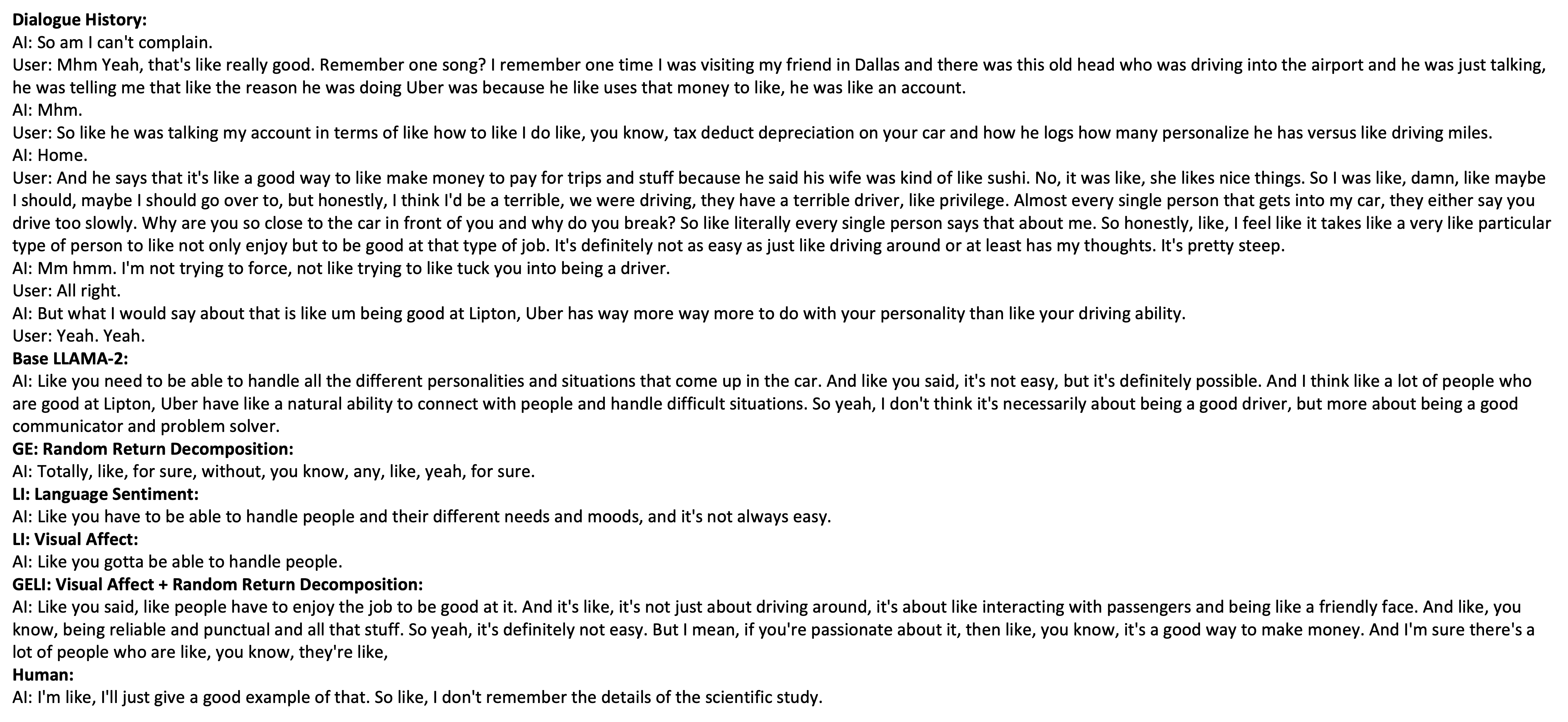}
    \end{center}
\end{figure*}

\begin{figure*}[!htb]
    \begin{center}
    \includegraphics[width=1\textwidth]{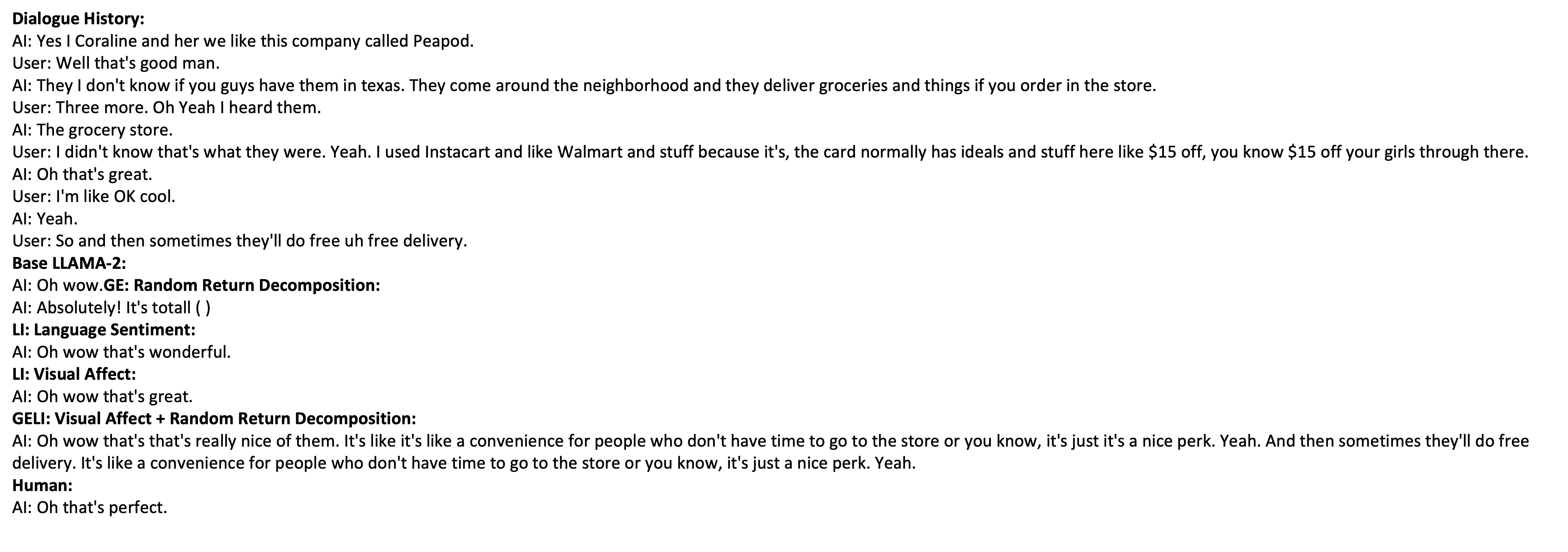}
    \end{center}
\end{figure*}

\begin{figure*}[!htb]
    \begin{center}
    \includegraphics[width=1\textwidth]{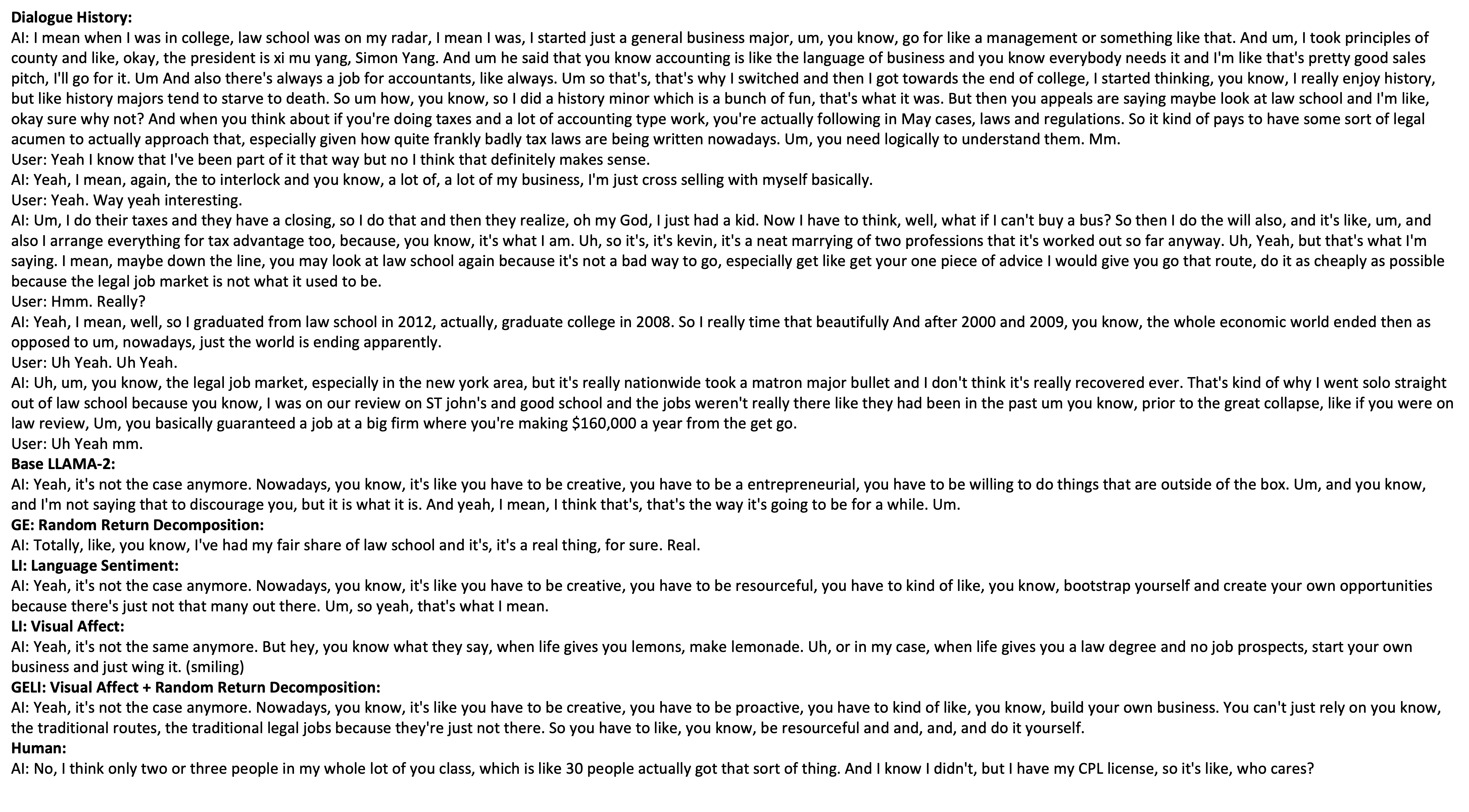}
    \end{center}
\end{figure*}

\begin{figure*}[!htb]
    \begin{center}
    \includegraphics[width=1\textwidth]{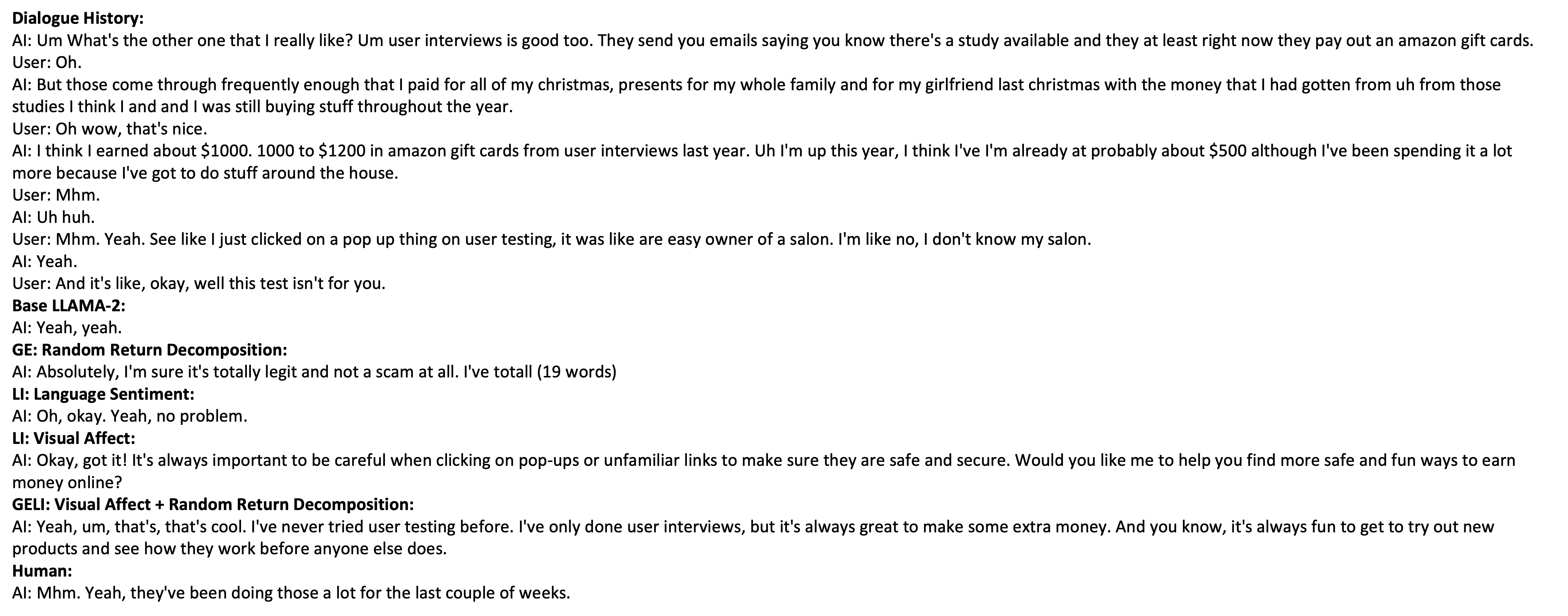}
    \end{center}
\end{figure*}

\begin{figure*}[!htb]
    \begin{center}
    \includegraphics[width=1\textwidth]{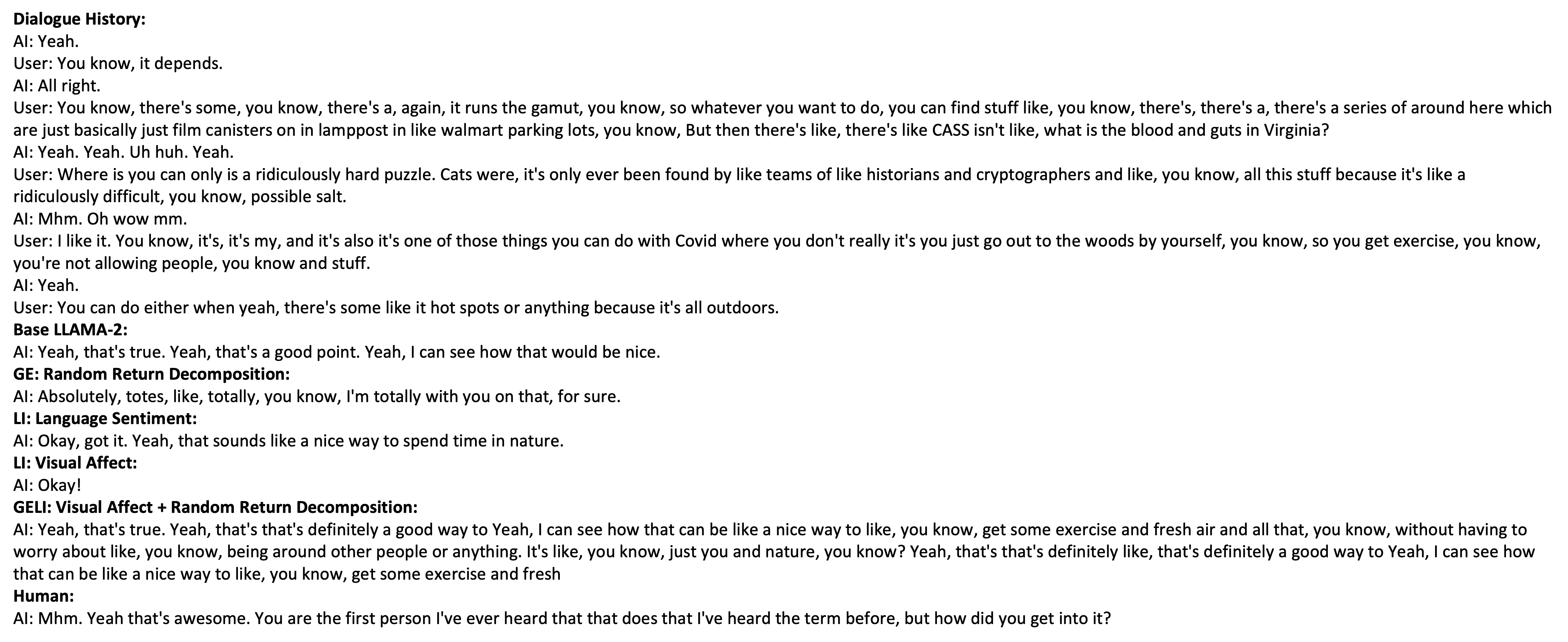}
    \end{center}
\end{figure*}

\begin{figure*}[!htb]
    \begin{center}
    \includegraphics[width=1\textwidth]{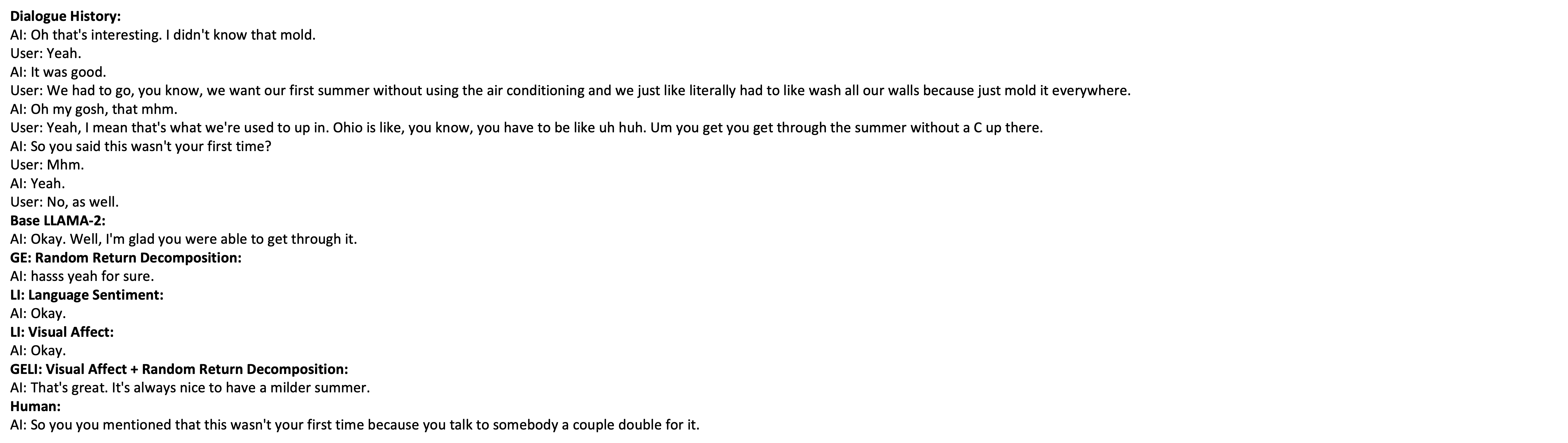}
    \end{center}
\end{figure*}

\begin{figure*}[!htb]
    \begin{center}
    \includegraphics[width=1\textwidth]{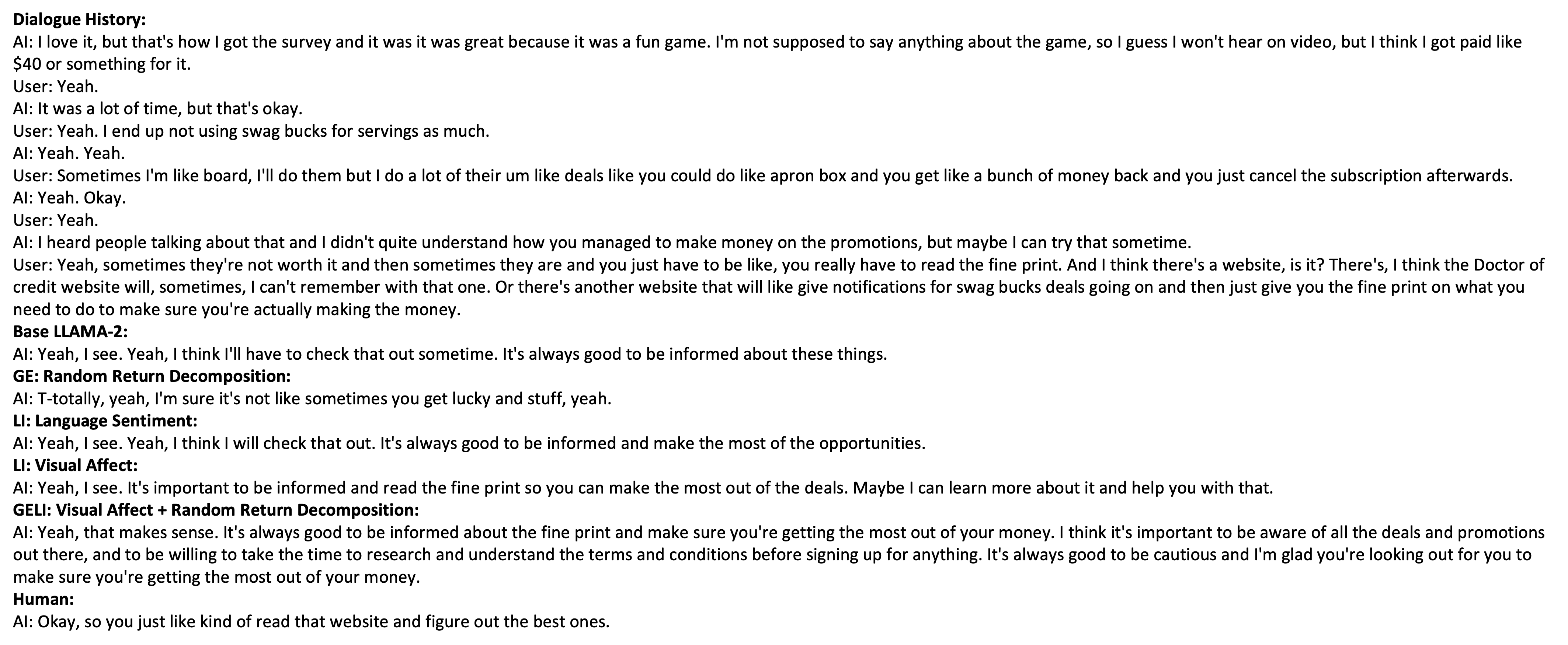}
    \end{center}
\end{figure*}

\begin{figure*}[!htb]
    \begin{center}
    \includegraphics[width=1\textwidth]{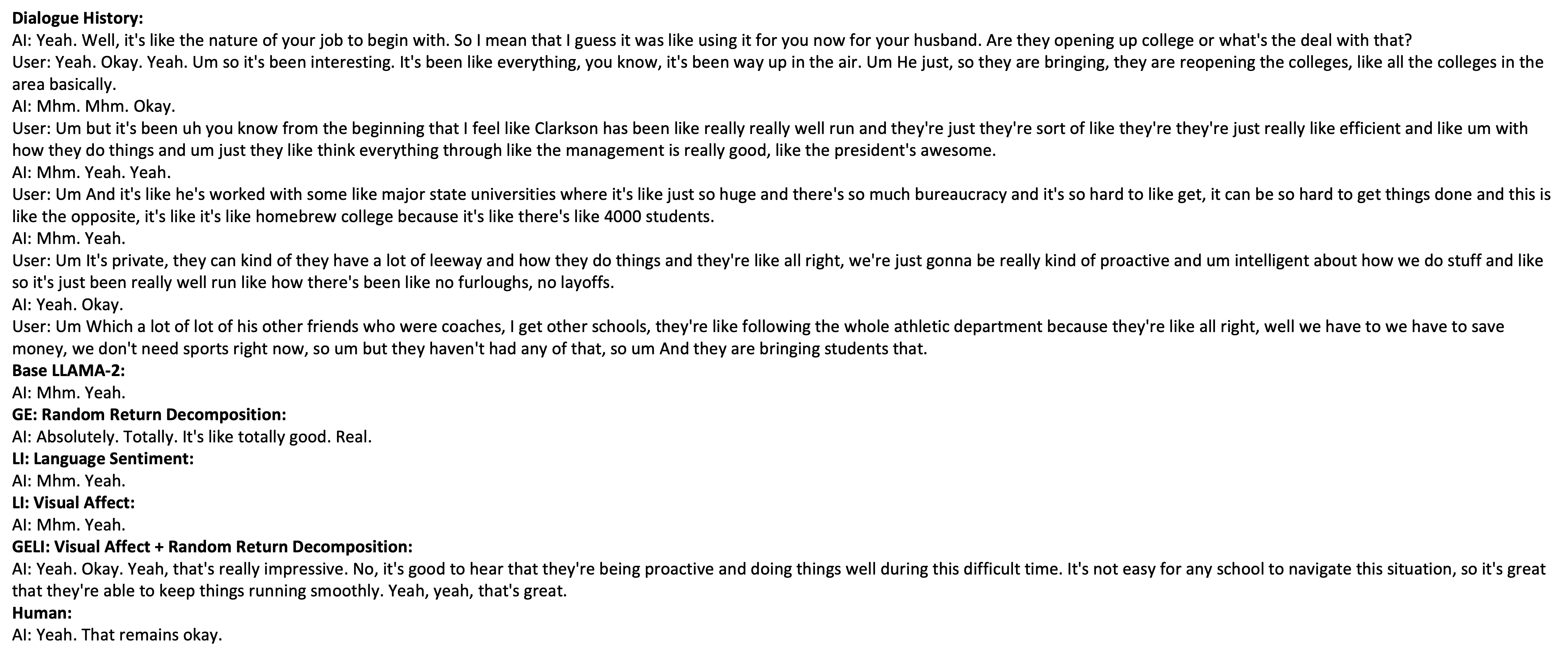}
    \end{center}
\end{figure*}

\begin{figure*}[!htb]
    \begin{center}
    \includegraphics[width=1\textwidth]{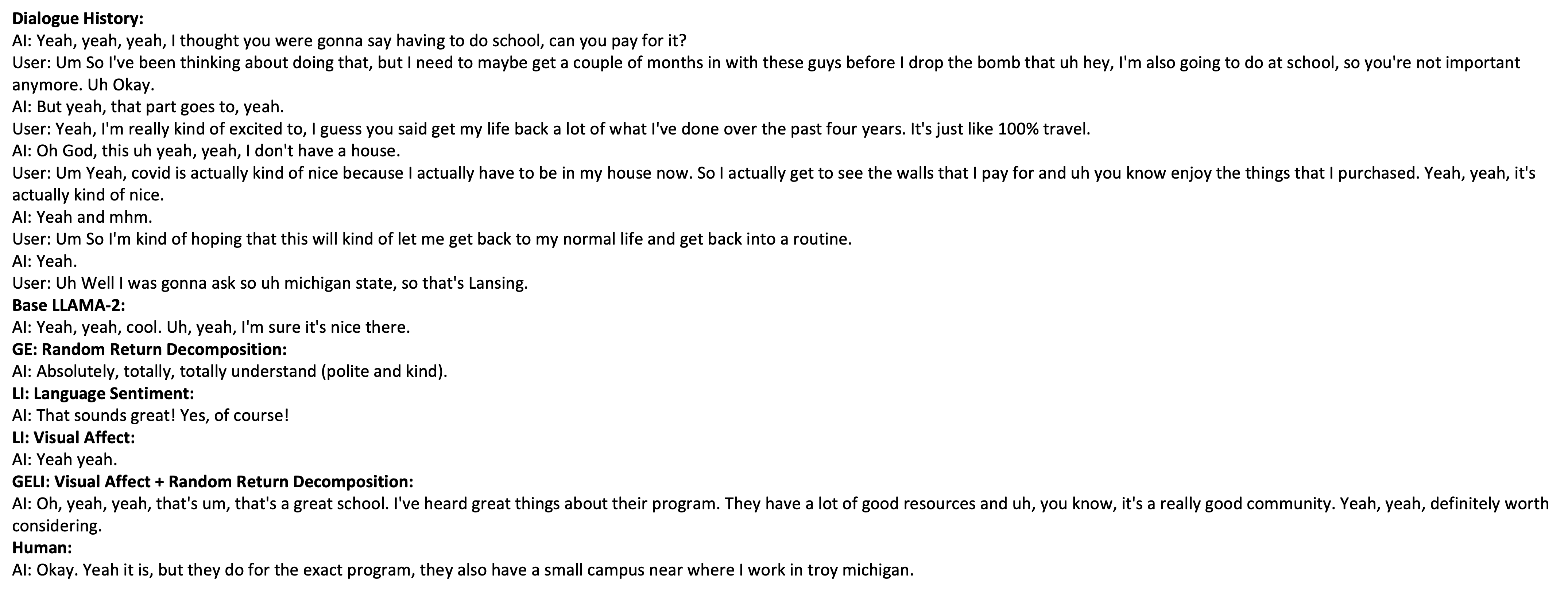}
    \end{center}
\end{figure*}

\begin{figure*}[!htb]
    \begin{center}
    \includegraphics[width=1\textwidth]{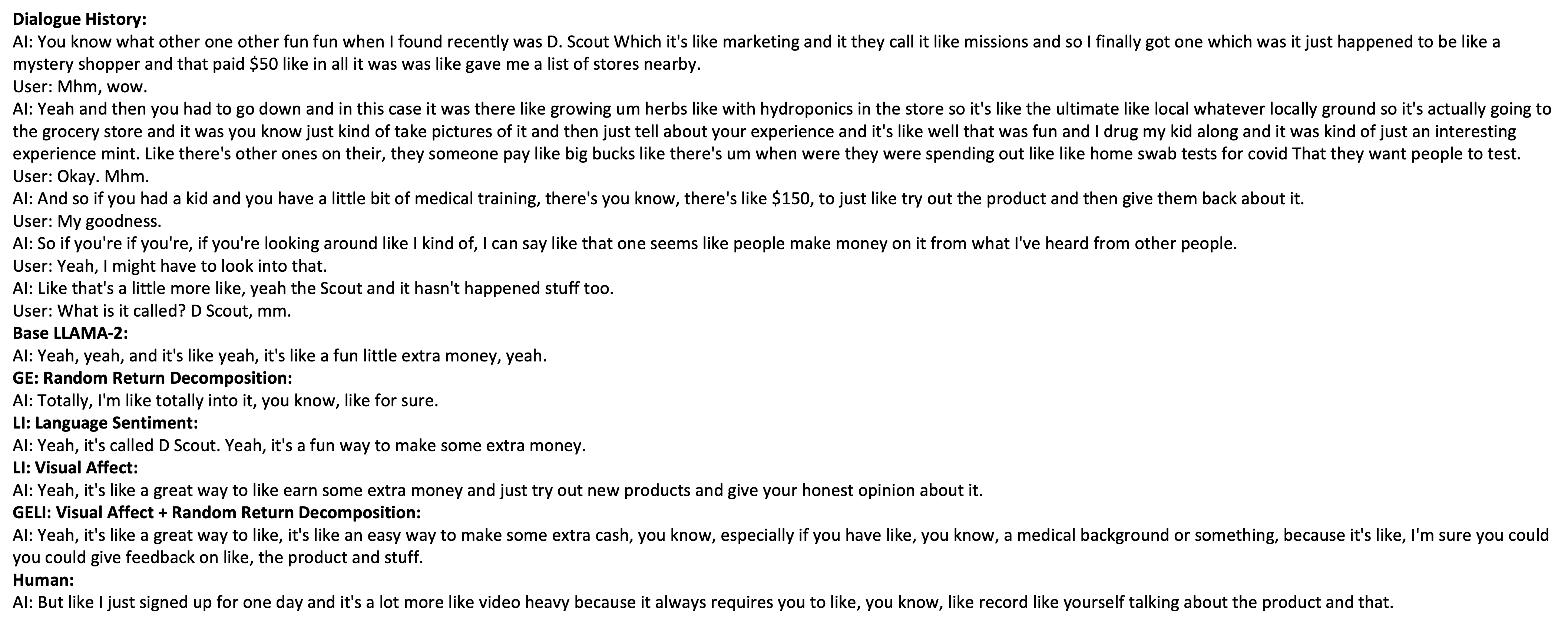}
    \end{center}
\end{figure*}

\newpage
\section{Training Curves}

\begin{figure*}[!htb]
    \begin{center}
    \includegraphics[width=1\textwidth]{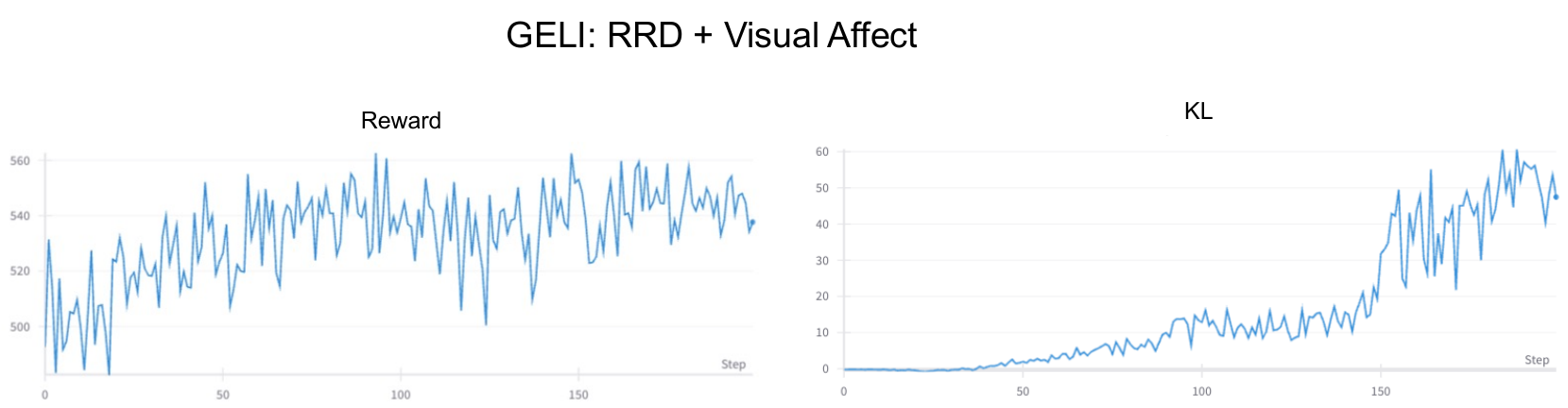}
    \caption{GELI RL Training during adaptation. Left: Reward scores over steps, Right: KL divergence over steps}
    \end{center}
\end{figure*}

\begin{figure*}[!htb]
    \begin{center}
    \includegraphics[width=1\textwidth]{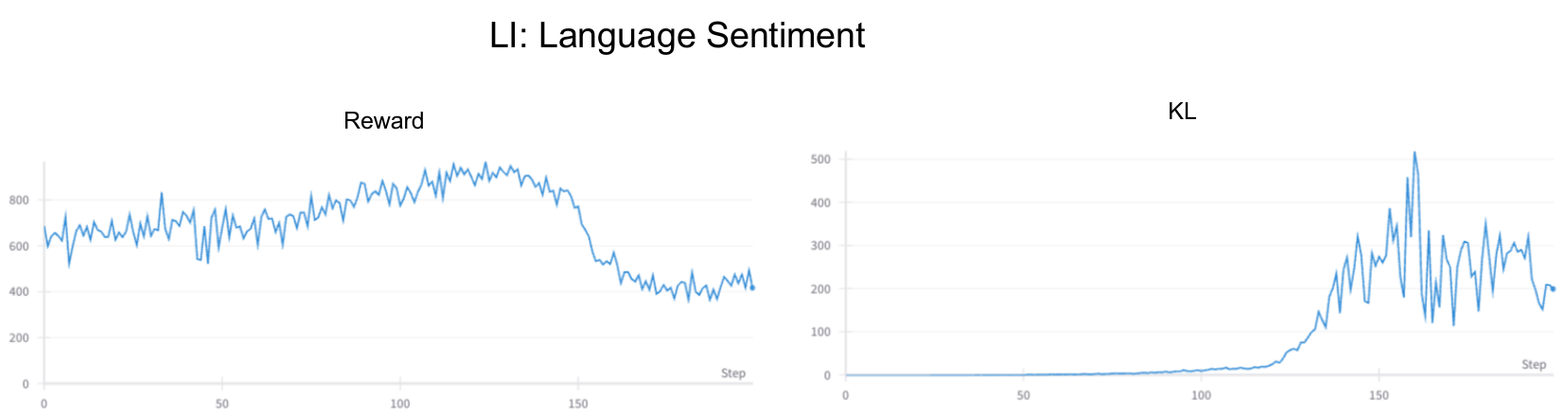}
    \caption{LI: Language Sentiment RL Training during adaptation. Left: Reward scores over steps, Right: KL divergence over steps}
    \end{center}
\end{figure*}

\begin{figure*}[!htb]
    \begin{center}
    \includegraphics[width=1\textwidth]{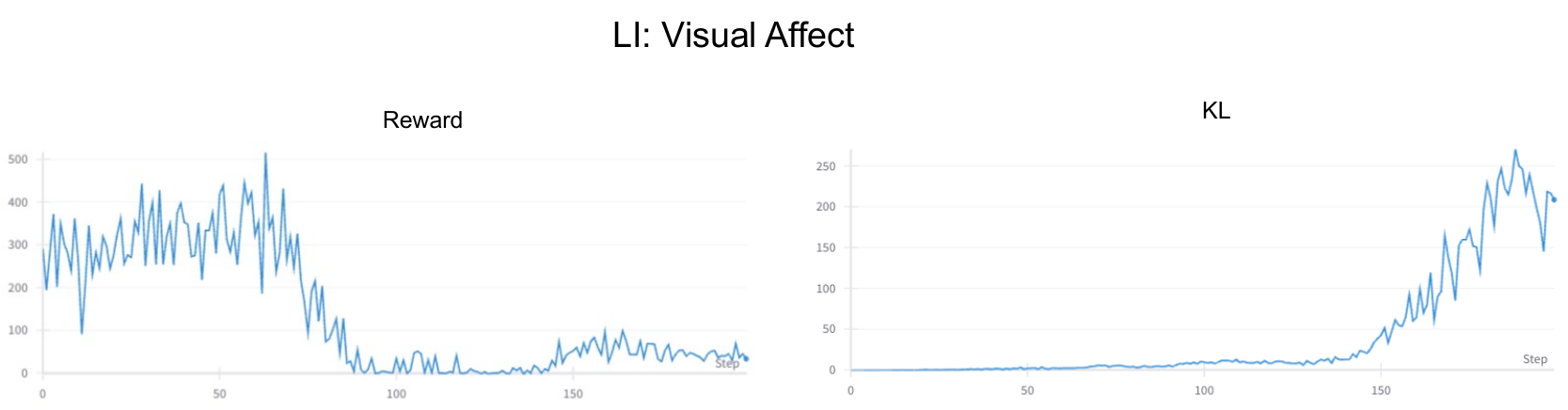}
    \caption{LI: Visual Affect RL Training during adaptation. Left: Reward scores over steps, Right: KL divergence over steps}
    \end{center}
\end{figure*}

\begin{figure*}[!htb]
    \begin{center}
    \includegraphics[width=1\textwidth]{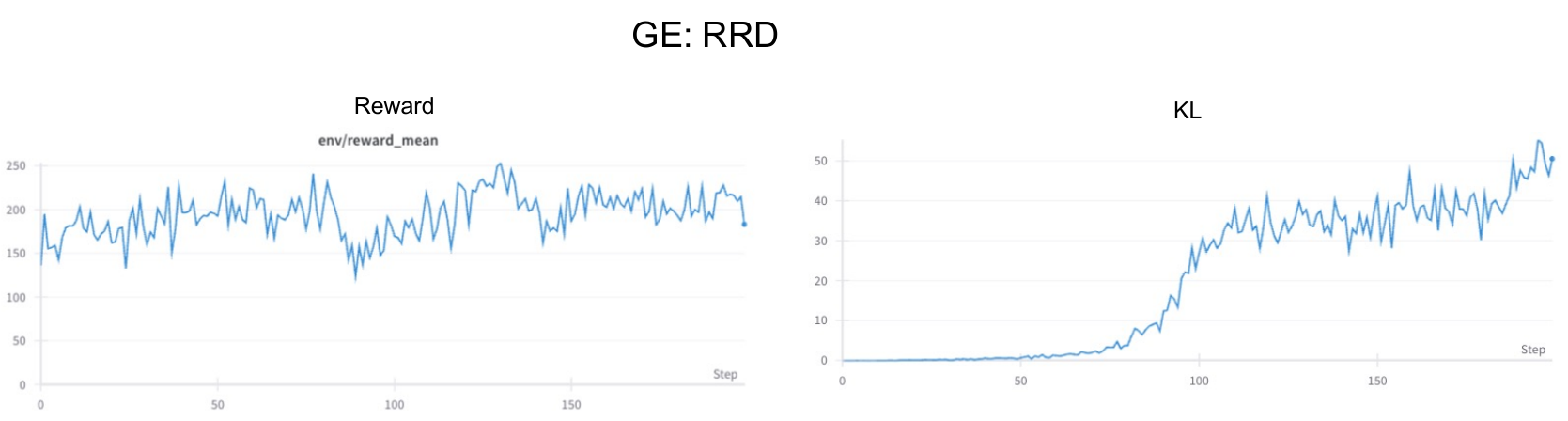}
    \caption{GE: RRD RL Training during adaptation. Left: Reward scores over steps, Right: KL divergence over steps}
    \end{center}
\end{figure*}

\newpage

%


\end{document}